\documentclass{article}
\usepackage{ijcai21}

\usepackage{times}

\usepackage{soul}
\usepackage{url}
\usepackage[hidelinks]{hyperref}
\usepackage[utf8]{inputenc}
\usepackage[small]{caption}
\usepackage{graphicx}
\usepackage{mathtools}
\usepackage{amsmath}
\usepackage{amsthm}
\usepackage{amssymb}

\usepackage{textcomp}

\usepackage{cuted}
\usepackage{multibib}
\newcites{SM}{Supplementary References}

\usepackage{tabularx,booktabs,ragged2e}
\newcolumntype{L}{>{\RaggedRight\arraybackslash}X}

\usepackage{physics}
\usepackage{bbm}

\usepackage{xurl}

\usepackage{listings}
\usepackage{xcolor}
\definecolor{codegreen}{rgb}{0,0.6,0}
\definecolor{codegray}{rgb}{0.5,0.5,0.5}
\definecolor{codepurple}{rgb}{0.58,0,0.82}
\definecolor{backcolour}{rgb}{0.95,0.95,0.92}

\lstdefinestyle{mystyle}{
    commentstyle=\color{codegreen},
    keywordstyle=\color{magenta},
    stringstyle=\color{codepurple},
    basicstyle=\ttfamily\footnotesize,
    breakatwhitespace=false,         
    breaklines=true,                 
    captionpos=b,                    
    keepspaces=true,                 
    showspaces=false,                
    showstringspaces=false,
    showtabs=false,                  
    tabsize=2
}

\newtheorem{theorem}{Theorem}
\newtheorem{lemma}{Lemma}
\newtheorem{claim}{Claim}
\newtheorem{definition}{Definition}

\makeatletter
\newif\if@restonecol
\makeatother

\usepackage[ruled,vlined]{algorithm2e}

\usepackage{todonotes}
\newcommand{\NEEDCITE}[1]{{\color{red} [CITATION NEEDED]}}

\title{Non-Parametric Stochastic Sequential Assignment With Random Arrival Times}

\author{
Danial Dervovic$^1$
\and
Parisa Hassanzadeh$^1$\and
Samuel Assefa$^{1}$\and
Prashant Reddy$^1$
\affiliations
$^1$J.P. Morgan AI Research\\
\emails
\{danial.dervovic, parisa.hassanzadeh, samuel.a.assefa, prashant.reddy\}@jpmorgan.com
}

\newenvironment{talign}
 {\align}
 {\endalign}
\newenvironment{talign*}
 {\csname align*\endcsname}
 {\endalign}

\begin{document}

\maketitle

\begin{abstract}
    We consider a problem wherein jobs arrive at random times and assume random values.
    Upon each job arrival, the decision-maker must decide immediately whether or not to accept the job and gain the value on offer as a reward, with the constraint that they may only accept at most $n$ jobs over some reference time period.
    The decision-maker only has access to $M$ independent realisations of the job arrival process. 
    We propose an algorithm, Non-Parametric Sequential Allocation (NPSA), for solving this problem. 
    Moreover, we prove that the expected reward returned by the NPSA algorithm converges in probability to optimality as $M$ grows large.
    We demonstrate the effectiveness of the algorithm empirically on synthetic data and on public fraud-detection datasets, from where the motivation for this work is derived.
\end{abstract}

\section{Introduction}

In industrial settings it is often the case that a positive class assignment by a classifier results in an expensive manual intervention.
A problem frequently arises whereby the number of these alerts exceeds the capacity of operators to manually investigate alerted examples~\cite{SiteReliabitlityEng}.
A common scenario is one where each example is further endowed with an intrinsic value along with its class label, with all negative examples having zero value to the operator.
Given their limited capacity, operators wish to maximise the cumulative value gained from expensive manual interventions.
As an example, in financial fraud detection~\cite{bolton2002}, this is manifested as truly fraudulent transactions having value to the operator as (some function of) the monetary value of the transaction, and non-fraudulent transactions yielding zero value~\cite{dal2015adaptive}.

In this paper we systematically account for the constraint on intervention capacity and desire to maximise reward, in the setting where selections are made in real-time and we have access to a large backlog of training data.
This problem structure is not limited to fraud, for example in cybersecurity~\cite{Vanek2012}, automated content moderation~\cite{CambridgeContentModeration}, compliance verification~\cite{avenhaus_canty_calogero_1996} and automated inspection in manufacturing~\cite{MORISHITA198359} there is a need for filtering a stream of comparable examples that are too numerous for exhaustive manual inspection, with the imperative of maximising the value of inspected examples.
We shall take an abstract view of \emph{jobs} arriving, each having an intrinsic value.

To this end, we extend a problem first considered by~\cite{Albright1974}, in which jobs arrive according to a random process and take on random nonnegative values.
At each job arrival, the decision-maker must decide immediately whether or not to accept the job and gain the value on offer as a reward.
They may only accept at most $n$ jobs over some reference time period.
In~\cite{Albright1974}, this problem is solved optimally by way of a system of ordinary differential equations (ODE). 
Importantly, the job arrival process is assumed to be known and admits a closed-form mathematical expression.
Solving the resulting system of ODEs analytically quickly becomes impractical, even for trivial job arrival processes.
We propose an efficient algorithm, \emph{Non-parametric Sequential Allocation Algorithm} (NPSA), which allows one merely to observe $M$ realisations of the job arrival process and still recover the optimal solution, as defined by this solution of ODEs, with high probability.
We empirically validate NPSA on both synthetic data and public fraud data, and rigorously prove its optimality.

This work plugs the gap in the literature where the following must be simultaneously accounted for: \emph{i.} explicit constraints on the number of job acceptances; \emph{ii.} maximising reward; \emph{iii.} treating job arrivals as a continuous-time random process; and \emph{iv.} learning the job value distribution and arrival process from data.

\paragraph{Related Work.}

The framework of \emph{Cost-sensitive learning}~\cite{Elkan2001} seeks to minimise the misclassification cost between positive and negative examples, even on an example-by-example basis~\cite{Bahnsen2014}, but often the methods are tuned to a specific classification algorithm and do not admit specification of an explicit constraint on the number of positive labels.
In~\cite{Shen2020}, the authors formulate fraud-detection as an RL problem. 
They explicitly take into account the capacity of inspections and costs, but operate in discrete-time and provide no theoretical guarantees.
Solving a \emph{Constrained MDP}~\cite{altman-constrainedMDP} optimises reward under long-term constraints that can be violated instantaneously but must be satisfied on average, such as in~\cite{DBLP:journals/corr/abs-2003-02189,DBLP:journals/corr/abs-2001-09377}.
Works such as~\cite{Mannor2006,pmlr-v48-jenatton16} on \emph{Constrained Online Learning} focus on a setting where the decision-maker interacts with an adversary and is simultaneously learning the adversary's behaviour and the best response, as measured by regret and variants thereof~\cite{pmlr-v108-zhao20b}.
Constraints typically relate to quantities averaged over sample paths~\cite{MannorShimkin2004}, whereas in our problem we have a discrete finite resource that is exhausted.
In this work we consider a non-adversarial environment that we learn before test-time from training data.
Moreover, the setting we focus on here explicitly is continuous-time and finite horizon, contrasting with the constrained MDP and online learning literature which considers discrete-time with an often infinite horizon.
Our problem aligns most closely with the framework of \emph{Stochastic Seqential Assignment Problems (SSAP)}~\cite{derman1972sequential,baharian2014stochastic}, where it is assumed that distributions of job values and the arrival process are known and closed-form optimal policies derived analytically; the question of learning from data is ignored~\cite{dupuis_wang_2002}.

\section{Problem Setup}\label{sec:setup}

We follow a modified version of the problem setup in~\cite{Albright1974}.
We assume a finite time horizon, from $t=0$ to $t=T$, over which jobs arrive according to a nonhomogeneous Poisson process with continuous intensity function $\lambda(t)$. 
There are a fixed number of indistinguishable workers, $n$, that we wish to assign to the stream of incoming jobs.
Each worker may only accept one job.
Every job has a nonnegative value associated to it that is gained as a reward by the decision-maker if accepted.
Any job that is not assigned immediately when it arrives can no longer be assigned.
It is assumed the total expected number of jobs that arrive over the horizon $[0, T]$ is much larger than the number of available workers, that is, $n \ll \int_0^T \lambda(t) \, \dd t$.

We take the job values to be i.i.d. nonnegative random variables drawn from a cumulative distribution $F$ with finite mean $0 < \mu < \infty$ and density $f$.
Moreover, we assume that the job value distribution is independent of the arrival process.
The decision-maker's goal is to maximise the total expected reward accorded to the $n$ workers over the time horizon $[0, T]$. 
We hereafter refer to Albright's problem as \textsf{SeqAlloc} (short for Sequential Allocation).


In the \textsf{SeqAlloc} model, it is assumed that $\lambda(t)$ and $F$ are known to the decision-maker ahead of time, and an optimal \emph{critical curve} $y_k(t)$ is derived for each of the $n$ workers.
When the $k$\textsuperscript{th} worker is active, if a job arrives at time $t$ with value greater than $y_k(t)$ the job is accepted, at which point the $(k-1)$\textsuperscript{th} worker is then active, until all $n$ workers have been exhausted.
These critical curves are addressed in more detail in Theorem~\ref{thm:albright}.

We modify the \textsf{SeqAlloc} problem setting in the following way.
The arrival intensity $\lambda(t)$ and $F$ are \emph{unknown} to the decision-maker ahead of time.
Instead, they have access to $M$ independent realisations of the job arrival process.
Each realisation consists of a list of tuples $(x_i, t_i)$, where $x_i$ is the reward for accepting job $i$ and $t_i$ its arrival time.
The goal for the decision-maker is the same as in the previous paragraph, that is, to derive critical curves for the $n$ workers so as to maximise the expected cumulative reward at test time.
In Section~\ref{sec:alg_main} we present an efficient algorithm for deriving these critical curves.
We hereafter refer to the modified problem we address in this paper as Non-Parametric \textsf{SeqAlloc}, or \textsf{SeqAlloc-NP} for short.

\section{Optimal Sequential Assignment}

Following~\cite{degroot1970,SAKAGUCHI1977KJ00001202133} we define a function that will take centre-stage in the sequel.
\begin{definition}[Mean shortage function]
\label{defn:mean_shortage}
For a nonnegative random variable $X$ with pdf $f$ and finite mean $\mu$, the \emph{mean shortage function} is given as
$\textstyle\phi(y) := \int^\infty_y (x - y) f(x)\, \dd x$ for $y \geq 0$.
\end{definition}

The next result follows from~\cite[Theorem 2]{Albright1974}.
\begin{theorem}[\textsf{SeqAlloc} critical curves]\label{thm:albright}
The (unique) optimal critical curves $y_n(t) \leq \ldots \leq y_1(t)$ solving the \textsf{SeqAlloc} problem satisfy the following system of ODEs (where $1 \leq k \leq n$):
\begin{align*}
\frac{\dd y_{k+1}(t)}{\dd t} = - \lambda(t) \left( \phi(y_{k+1}(t)) - \phi(y_{k}(t)) \right), \\ \phi(y_{0}(t)) =0, \qq{} y_k(T) = 0,\qq{} t \in [0, T].
\end{align*}
\end{theorem}
Indeed, solving this system of ODEs exactly is generally intractable, as we shall see in more detail in Section~\ref{sec:expt}.
Theorem~\ref{thm:albright} provides the optimal solution to the \textsf{SeqAlloc} problem.

\subsection{Numerical Algorithm for \textsf{SeqAlloc-NP}}\label{sec:alg_main}

An algorithm to solve the non-parametric problem, \textsf{SeqAlloc-NP}, immediately suggests itself as shown in Algorithm~\ref{alg:main_meta}: use the $M$ independent realisations of the job arrival process to approximate the intensity $\lambda(t)$ and the mean shortage function $\phi(y)$, then use a numerical ODE solver with Theorem~\ref{thm:albright} to extract critical curves.

\begin{algorithm}
\DontPrintSemicolon
\caption{NPSA: solution to \textsf{SeqAlloc-NP}}
\label{alg:main_meta}
\SetKwInOut{Input}{Input}
\Input{Number of workers $n$, ODE solver $\mathcal{D}$}
\KwData{$M$ realisations of job arrival process, $\mathcal{M}$}
\SetKwInOut{Output}{Output}
\Output{Critical curves $\{\widetilde{y}_k(t)\}_{k=1}^n$}
\BlankLine
\Begin{
    Estimate $\widetilde{\lambda}(t)$ and $\widetilde{\phi}(y)$ from $\mathcal{M}$ \;
    $\widetilde{y}_0(t) \leftarrow \infty$, $Y \leftarrow \{\widetilde{y}_0(t)\}$\;
    \For{$k$ \KwSty{in} $(1, \ldots, n)$}{
        Solve via $\mathcal{D}$: $\widetilde{y}_k(T) = 0, t \in [0, T]$.  \; $\frac{\operatorname{d}\! \widetilde{y}_{k}(t)}{\operatorname{d}\!t} = - \widetilde{\lambda}(t) \left( \widetilde{\phi}(\widetilde{y}_{k}(t)) - \widetilde{\phi}(\widetilde{y}_{k-1}(t)) \right),$\;
        
        $Y \leftarrow Y \cup \{\widetilde{y}_k(t)\}$
    }
    \KwRet{$Y \setminus \widetilde{y}_0(t)$}
}
\end{algorithm}

Algorithm~\ref{alg:main_meta} is a meta-algorithm in the sense that the estimators $\widetilde{\lambda}(t)$ and $\widetilde{\phi}$ must be defined for a full specification.
These estimators must be accurate so as to give the correct solution and be efficient to evaluate, as the numerical ODE solver will call these functions many times.
In Section~\ref{sec:est_non_hom_Poisson} we define $\widetilde{\lambda}(t)$ and in Section~\ref{sec:phi_via_eccdf} we define $\widetilde{\phi}$ appropriately.
Taken together with Algorithm~\ref{alg:main_meta} this defines the \emph{Non-parametric Sequential Allocation Algorithm}, which we designate by NPSA for the remainder of the paper.

\subsection{Estimation of Non-Homogeneous Poisson Processes}\label{sec:est_non_hom_Poisson}

In this section we discuss estimation of the non-homogeneous Poisson process $P$ with rate function $\lambda(t) > 0$ for all $t \in [0, T]$.
We make the assumption that we have $M$ i.i.d. observed realisations of this process.
In this case, we adopt the well known technique of~\cite{LawKeltonBook} specialised by~\cite{HENDERSON2003375}. 
Briefly, the rate function estimator is taken to be piecewise constant, with breakpoints spaced equally according to some fixed width $\delta$.

Denote by $\widetilde{\lambda}^{(M)}(t)$ the estimator of $\lambda(t)$ by $M$ independent realisations of $P$.
Let the subinterval width used by the estimator be $\delta_M > 0$.
We denote by $C_i(a, b)$ the number of jobs arriving in the interval $[a, b)$ in the $i$\textsuperscript{th} independent realisation of $P$.
For $t \geq 0$, let $\ell(t) := \left\lfloor t / \delta_M \right\rfloor \cdot \delta_M$
so that $t \in [\ell(t), \ell(t) + \delta_M]$.
Our estimator is the number of arrivals recorded within a given subinterval, averaged over independent realisations of $P$ and normalised by the binwidth $\delta_M$, that is,
\begin{equation}\label{eq:rate_estimator}
    \widetilde{\lambda}^{(M)}(t) = \frac{1}{M \delta_M} \sum_{i = 1}^M C_i(\ell(t), \ell(t) + \delta_M).
\end{equation}
From~\cite[Remark 2]{HENDERSON2003375} we have the following result.
\begin{theorem}[Arrival rate estimator convergence]\label{thm:rate_estimator_convergence}
Suppose that $\delta_M = O(M^{-a})$ for any $a \in (0, 1)$ and fix $t \in [0, T)$.
Then, $\widetilde{\lambda}^{(M)}(t) \to \lambda(t)$ almost surely as $M \to \infty$.
\end{theorem}

For the NPSA algorithm we use Eq.~\eqref{eq:rate_estimator} with $\delta_M = T \cdot M^{-\frac{1}{3}}$ as the estimator for the intensity $\lambda(t)$.
There are $\left\lceil T / \delta_M \right\rceil$ ordered subintervals, so the time complexity of evaluating $\widetilde{\lambda}(t)$ is $O(\log (T/\delta_M ) )$ and the space complexity is $O(T / \delta_M)$, owing respectively to searching for the correct subinterval $[\ell(t), \ell(t) + \delta_M]$ via binary search and storing the binned counts $C_i$ .
The initial computation of the $C_i$ incurs a time cost of $O(M N_{\text{max}})$, where $N_{\text{max}}$ denotes the maximum number of jobs over the $M$ realisations.

\subsection{Mean Shortage Function Estimator}
\label{sec:phi_via_eccdf}

The following result (with proof in the Supplementary Material) leads us to the $\widetilde{\phi}$ estimator for NPSA.

\begin{lemma}\label{lem:mean_shortage_integral}
The mean shortage function of Definition~\ref{defn:mean_shortage} can be written as $\textstyle\phi(y) = \int^\infty_y (1 - F(x)) \dd x$, 
where $F$ is the cdf of the random variable $X$.
\end{lemma}
Lemma~\ref{lem:mean_shortage_integral} suggests the following estimator: perform the integral in Lemma~\ref{lem:mean_shortage_integral}, replacing the cdf $F$ with the \emph{empirical cdf} for the job value r.v. $X$ computed with the samples $(x_1, \ldots, x_N)$, $F_N(x) := \frac{1}{N} \sum_{i=1}^N \vb*{1}_{x_i \leq x}$, where $\vb*{1}_{\omega}$ is the indicator variable for an event $\omega$.
Since the empirical cdf is piecewise constant, the integral is given by the sum of areas of $O(N)$ rectangles.
Concretely, we cache the integral values $\phi_i$ evaluated at each data sample $x_i$ and linearly interpolate for intermediate $y \in [x_i, x_{i+1})$ at evaluation time.
Indeed, after initial one-time preprocessing, this estimate $\widetilde{\phi}_N(y)$ has a runtime complexity of $O(\log N)$ per function call (arising from a binary search of the precomputed values) and space complexity $O(N)$, where $N$ is the number of data samples used for estimation.
Pseudocode for these computations is given in the Supplementary Material.


We have shown that the NPSA mean-shortage function estimator is computationally efficient.
It now remains to show that it is accurate, that is, statistically consistent.

\begin{theorem}\label{thm:ecdf_convergence}
Let $X$ be a nonnegative random variable with associated mean shortage function $\phi$.
Then, the estimate of the mean shortage function converges in probability to the true value, that is,
$$\lim_{N \to \infty} \mathbb{P} \qty[ \sup_{y \geq 0} \abs{\widetilde{\phi}_{N}(y) - \phi(y)} > \epsilon] = 0$$
for any $\epsilon > 0$, where the estimate computed by the estimator using $N$ independent samples of $X$ is denoted by $\widetilde{\phi}_{N}(y)$.
\end{theorem}
\begin{proof}[Proof Sketch]
It can be shown that an upper-bound on $\abs*{\widetilde{\phi}_N(y) - \phi(y)}$ is induced by an upper-bound on $\abs{F_N(x) - F(x)}$.
The Dvoretzky–Kiefer–Wolfowitz inequality~\cite{dvoretzky1956,massart1990} furnished with this bound yields the result.
\end{proof}

\subsection{NPSA Performance Bounds}

We have shown that the individual components of the NPSA algorithm, namely the intensity $\widetilde{\lambda}(t)$ and mean shortage $\widetilde{\phi}(y)$ estimators, are computationally efficient and statistically consistent.
However, our main interest is in the output of the overall NPSA algorithm, that is, will following the derived threshold curves at test time yield an expected reward that is optimal with high probability?
The answer to this question is affirmative under the assumptions of the \textsf{SeqAlloc-NP} problem setup as described in Section~\ref{sec:setup}.

We will need some results on approximation of ODEs.
Following the presentation of~\cite{Brauer1963}, consider the initial value problem
\begin{equation}\label{eq:ivp}
    \frac{\dd x}{\dd t} = f(t, x),
\end{equation}
where $x$ and $f$ are $d$-dimensional vectors and $0 \leq t < \infty$.
Assume that $f(t, x)$ is continuous for $0 \leq t < \infty$, $\norm{x} < \infty$ and $\norm{\,\cdot\,}$ is a norm.
Recall that a continuous function $x(t)$ is an \emph{$\epsilon$-approximation} to~\eqref{eq:ivp} for some $\epsilon \geq 0$ on an interval if it is differentiable on an interval $I$ apart for a finite set of points $S$, and $\norm*{\frac{\dd x(t)}{\dd t} - f(t, x(t))} \leq \epsilon$ on $I \setminus S$.
The function $f(t, x)$ satisfies a \emph{Lipschitz condition} with constant $L_f$ on a region $D \subset \mathbb{R} \times \mathbb{R}^d$ if $\norm*{f(t, x) - f(t, x')} \leq L_f \norm{x - x'}$
whenever $(t, x), (t, x') \in D$.
We will require the following lemma from~\cite{Brauer1963}.
\begin{lemma}\label{lem:eps_approx_dist}
Suppose that $x(t)$ is a solution to the initial value problem~\eqref{eq:ivp} and $x'(t)$ is an $\epsilon$-approximate solution to $\eqref{eq:ivp}$. 
Then
\begin{equation*}\textstyle
    \norm{x(t) - x'(t)} \leq \norm{x(0) - x'(0)}e^{L_f t} + \frac{\epsilon}{L_f}(e^{L_f t } - 1),
\end{equation*}
where $L_f$ is the Lipschitz-constant of $f(t, x)$.
\end{lemma}

Now consider two instantiations of the problem setup with differing parameters, which we call \emph{scenarios}: one in which the job values are nonnegative r.v.s $X$ with mean $\mu$, cdf $F$ and mean shortage function $\phi$; in the other, the job values are nonnegative r.v.s $X'$ with mean $\mu'$, cdf $F'$ and mean shortage function $\phi'$.
We stipulate that $X$ and $X'$ have the same support and admit the (bounded) densities $f$ and $f'$ respectively.
In the first scenario the jobs arrive with intensity function $\lambda(t) > 0$ and in the second they arrive with intensity $\lambda'(t) > 0$.
In both scenarios there are $n$ workers.
We are to use the preceding results to show that the difference between threshold curves $\abs{y_k(t) -  y'_k(t)}$ computed between these two scenarios via NPSA can be bounded by a function of the scenario parameters.

We further stipulate that the scenarios do not differ by too great a degree, that is, $(1 - \delta_\lambda) \lambda(t) \leq \lambda'(t) \leq (1 + \delta_\lambda) \lambda(t)$ for all $t \in [0, T]$ and $(1 - \delta_\phi) \phi(y) \leq \phi'(y) \leq (1 + \delta_\phi) \phi(y)$ for all $y \in [0, \infty)$, where $0 < \delta_\lambda, \delta_\phi < 1$.
Moreover, define $\lambda_{\text{max}} = \max_{t \in [0, T]} \qty{ \lambda(t) }$.
We are led to the following result.
\begin{lemma}\label{lem:y_prime_epsilon_approx}
For any $k \in \{1, \ldots, n\}$, $y'_k(t)$ is an $\epsilon$-approximator for $y_k(t)$ when $\epsilon > 2 \mu \lambda_{\text{max}} (\delta_\phi + \delta_\lambda)$.
\end{lemma}
\begin{proof}[Proof Sketch]
Upper bound the left-hand-side of
$$\abs{\lambda'(t)\qty( \phi'(y'_{k+1}) - \phi'(y'_k) ) - \lambda(t)\qty( \phi(y'_{k+1}) - \phi(y'_k) ) } < \epsilon,$$
where the ODE description of $y_k$ and $y'_k$ is employed from Theorem~\ref{thm:albright} and use the definition of an $\epsilon$-approximator.
\end{proof}

We are now able to compute a general bound on the difference between threshold curves derived from slightly differing scenarios.

\begin{lemma}\label{lem:threshold_dist_bound}
    For any $k \in \{1, \ldots , n\}$
    \begin{equation*}
        \abs{y_k(t) - y'_k(t)} \leq (\delta_\lambda + \delta_\phi) \mu \qty( e^{2 \lambda_{\text{max}} (T - t) } - 1 ).  
    \end{equation*}
\end{lemma}
\begin{proof}[Proof Sketch]
Compute the Lipschitz constant $2\lambda_{\text{max}}$ for the ODE system of Theorem~\ref{thm:albright}, then use Lemmas~\ref{lem:eps_approx_dist}~and~\ref{lem:y_prime_epsilon_approx}.
\end{proof}

Having bounded the difference between threshold curves in two differing scenarios, it remains to translate this difference into a difference in reward.
Define
\begin{equation}
   H(y) := \textstyle\int^\infty_y x f(x) \dd x ; \qq{} \overline{F}(y) := 1 - F(y),
\end{equation}
so that $\phi(y) = H(y) - y \overline{F}(y)$ by Lemma~\ref{lem:mean_shortage_integral}.
We also have from~\cite[Theorem 2]{Albright1974} that for a set of (not necessarily optimal) threshold curves $\{\widetilde{y}_k(t) \}_{k=1}^n$, the expected reward to be gained by replaying the thresholds from a time $t\in [0, T]$ is given by 
\begin{align}
    \label{eq:expected_reward_integral}
    &E_k(t ; \widetilde{y}_k, \ldots, \widetilde{y}_1) = \nonumber \\
    &\quad\int_t^T \bigg[ H(\widetilde{y}_k(\tau)) +  \overline{F}(\widetilde{y}_k(\tau))\cdot E_{k-1}(\tau ; \widetilde{y}_{k-1}, \ldots, \widetilde{y}_1) \bigg] \nonumber\\
    &\quad\hphantom{=\int_t^T} \times \bigg[ \lambda(\tau)\exp[- \textstyle\int^\tau_t \lambda(\sigma) \overline{F}(\widetilde{y}_k(\sigma)) \dd \sigma ]\bigg] \dd \tau.
\end{align}
We also have from~\cite[Theorem 1]{Albright1974} that
\begin{align}
    \label{eq:expected_reward_thresholds}\textstyle
    E_n(t ; \widetilde{y}_n, \ldots, \widetilde{y}_1) = \sum_{k=1}^n \widetilde{y}_k(t)
\end{align}

We wish to lower-bound the expected total reward at test time using the thresholds $\{y'_k\}_{k=1}^n$, when the job arrival process has value distribution $F$ and intensity $\lambda(t)$.

We first make the assumptions that the functions $H$ and $\overline{F}$ do not differ too greatly between the critical curves derived for the two scenarios.
Concretely, there exist $\epsilon_{\overline{F}}, \delta_{\overline{F}}, \delta_{H} \in (0, 1)$ such that $e^{\delta_H} H(y_k(t)) \geq H(y'_k(t)) \geq e^{-\delta_H} H(y_k(t))$, $e^{\delta_{\overline{F}}}\overline{F}(y_k(t)) \geq \overline{F}(y'_k(t)) \geq e^{-\delta_{\overline{F}}} \overline{F}(y_k(t))$ and $\abs{\overline{F}(y'_k(t)) - \overline{F}(y_k(t))} \leq \epsilon_{\overline{F}}$ for all $k \in \qty{1, \ldots, n}$ and $t \in [0, T]$.
Furthermore, we define the mean arrival rate $\overline{\lambda} := \frac{1}{T} \int_0^T \lambda(t) \dd t$.
We are then able to prove the following lower bound on the total reward under incorrectly specified critical curves.

\begin{lemma}\label{lem:reward_lower_bound}
Let $\delta = \max\qty{ \delta_{H}, \delta_{\overline{F}}}$.
Then
\begin{equation*}
    E_n (t ; y'_k, \ldots, y'_1) \geq e^{-n(\delta + \overline{\lambda}\epsilon_{\overline{F}}T)} E_n(t ; y_n, \ldots, y_1).
\end{equation*}
\end{lemma}
\begin{proof}[Proof Sketch]
Use induction on $n$ and Eq.~\eqref{eq:expected_reward_integral}.
\end{proof}

We have the ingredients to prove the main result.
There is a technical difficulty to be overcome, whereby we need to push additive errors from Lemma~\ref{lem:threshold_dist_bound} through to the multiplicative errors required by Lemma~\ref{lem:reward_lower_bound}. 
This is possible when the functions $H$, $\overline{F}$ and $\phi$ are all Lipschitz continuous and have positive lower-bound on $\bigcup_{k=1}^n \operatorname{Range}\qty(y_k(t)) \cup \operatorname{Range}\qty(y'_k(t))$, facts which are established rigorously in the Supplementary Material.

As a shorthand, we denote the expected reward gained by using the optimal critical curves by $r^\star := E_n(0; y_n, \ldots, y_1)$.
Moreover, let the critical curves computed by NPSA from $M$ job arrival process realisations be $\{ \widetilde{y}_k^{(M)} \}_{k=1}^n$ and the associated expected reward under the true data distribution be the random variable $R^{(M)} := E_n(0; \widetilde{y}_n^{(M)}, \ldots, \widetilde{y}_1^{(M)})$.

\begin{theorem}\label{thm:optimal_convergence}
Fix an arbitrary $\epsilon \in (0,1)$.
Then,
$$\lim_{M \to \infty}\mathbb{P}\qty[ \frac{R^{(M)}}{r^\star} \geq 1-\epsilon ] = 1.$$
\end{theorem}
\begin{proof}[Proof Sketch]
Using Lemma~\ref{lem:reward_lower_bound} one can lower bound the probability by 
\begin{equation}\label{eq:thm:optimal_convergence_sketch_1}\textstyle
    1 -  \mathbb{P}\qty[\delta_H > \frac{2\epsilon}{n}] - \mathbb{P}\qty[\delta_{\overline{F}} > \frac{2\epsilon}{n}] - \mathbb{P}\qty[\epsilon_{\overline{F}} > \frac{2\epsilon }{n \overline{\lambda}T}].
\end{equation}
Lemma~\ref{lem:threshold_dist_bound} and Theorem~\ref{thm:ecdf_convergence} can be used to show that the third term of~\eqref{eq:thm:optimal_convergence_sketch_1} vanishes as $M \to \infty$.
The first and second term also vanish by pushing through the additive error from Theorem~\ref{thm:ecdf_convergence}~and~Lemma~\ref{lem:threshold_dist_bound} into multiplicative errors, then verifying these quantities are small enough when $M$ is sufficiently large. 
\end{proof}

Theorem~\ref{thm:optimal_convergence} demonstrates that the NPSA algorithm solves the \textsf{SeqAlloc-NP} problem optimally when the number of realisations of the job arrival process, $M$, is sufficiently large.

\begin{figure}
    \centering
    \includegraphics[width=0.49\columnwidth]{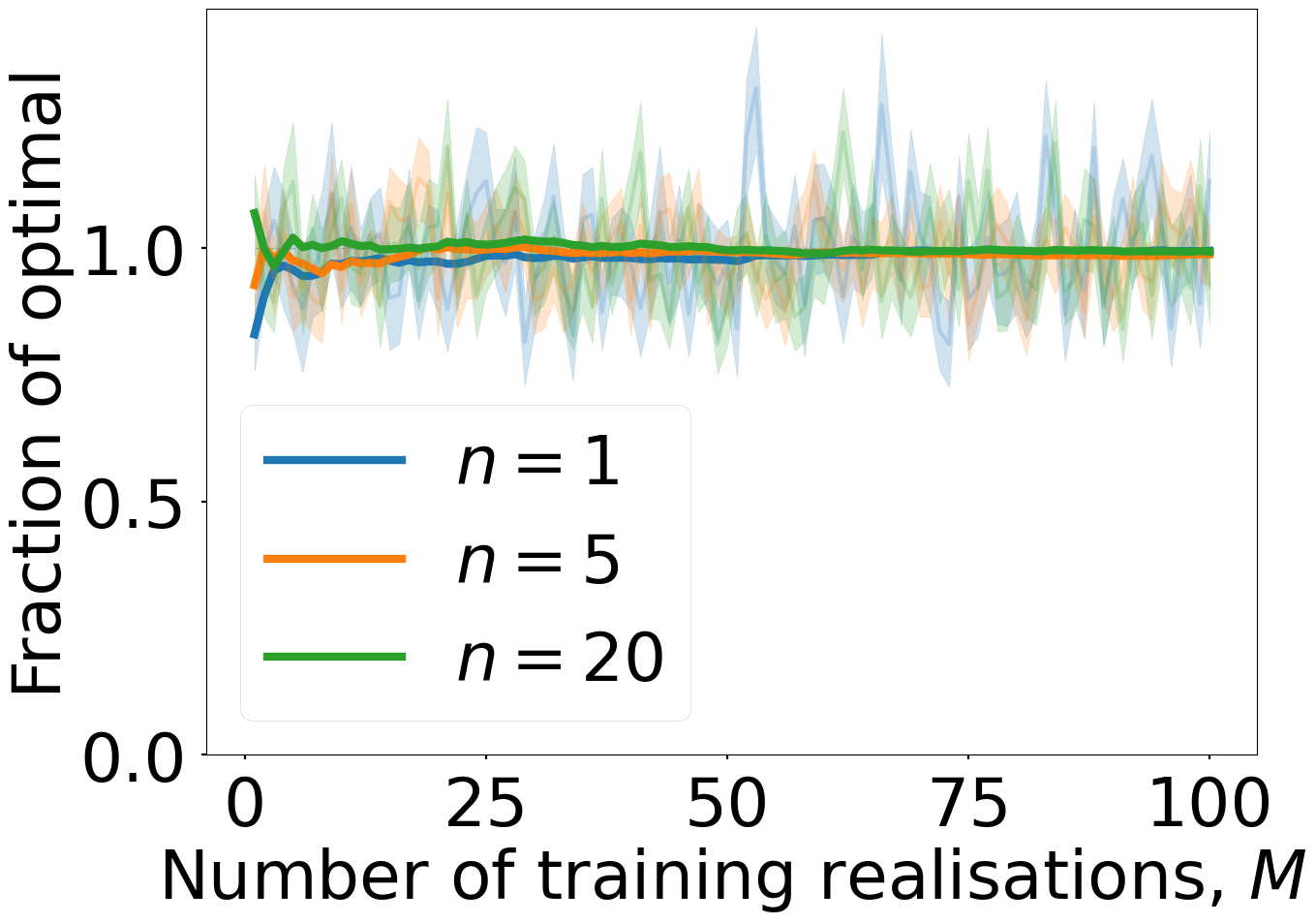}
    \includegraphics[width=0.48\columnwidth]{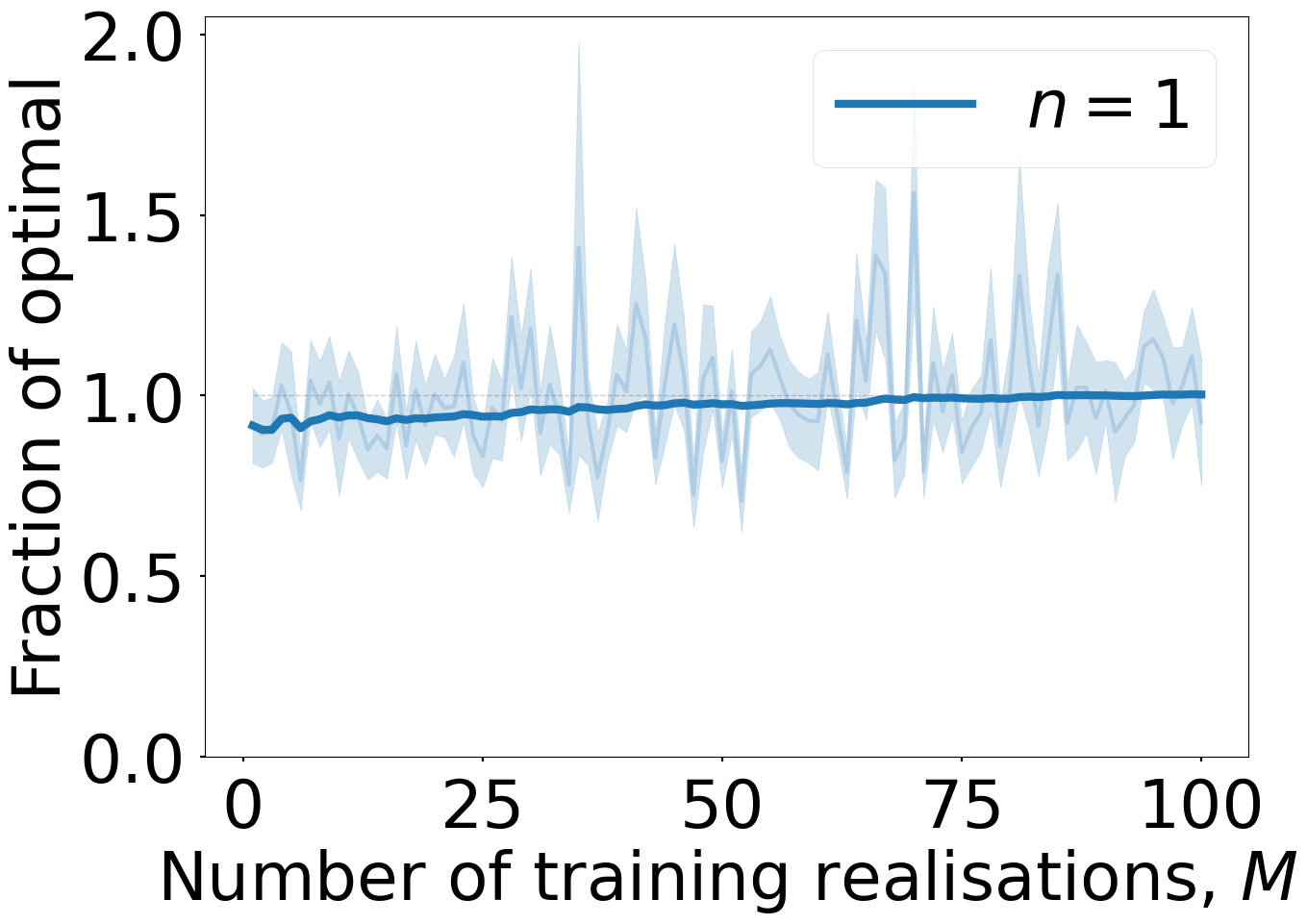}
    \caption{Convergence experiments for job values that are exponentially distributed (left) with mean $\mu = 5$ and job values that are Lomax-distributed (right) with shape $\alpha=3.5$ and scale $\xi=5$.
    The time horizon $T = 2\pi$ and the job arrival rate is $\lambda = 1$.
    The horizontal dashed line at $y = 1$ indicates optimal reward.
    The rolling (Cesàro) average is drawn with thick lines to highlight convergence.
    }
    \label{fig:varying_M_expt}
\end{figure}

\section{Experiments}\label{sec:expt}

We now empirically validate the efficacy of the NPSA algorithm.
Three experiments are conducted: \emph{i.} observing the convergence of NPSA to optimality; \emph{ii.} assessing the impact on NPSA performance when the job value distribution $F$ and the arrival intensity $\lambda(t)$ of the data-generating process differ between training and test time; and finally \emph{iii.} applying NPSA to public fraud data and evaluating its effectiveness in detection of the most valuable fraudulent transactions.

\begin{figure}
    \centering
    \includegraphics[width=0.47\columnwidth]{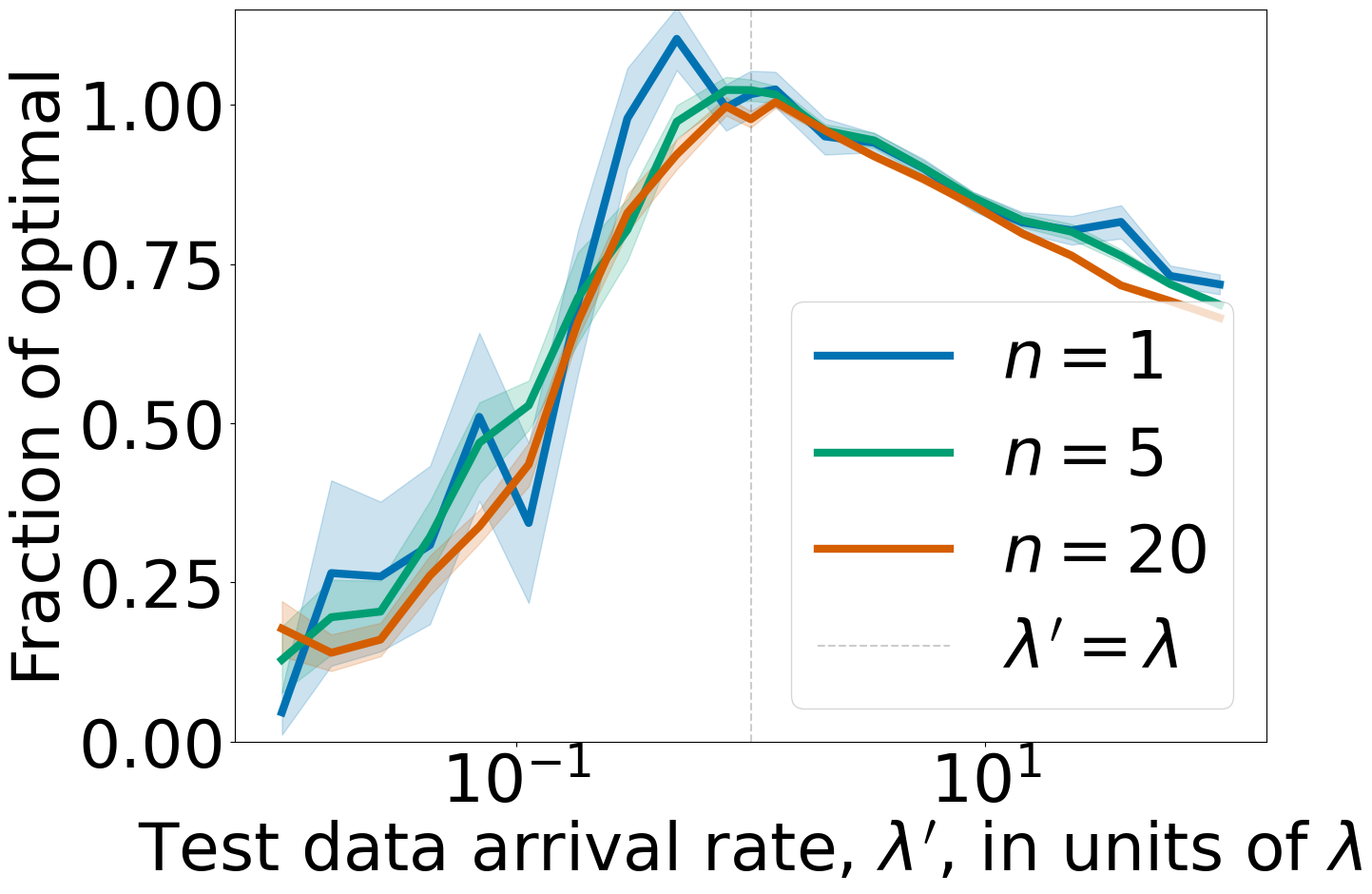}
    \includegraphics[width=0.49\columnwidth]{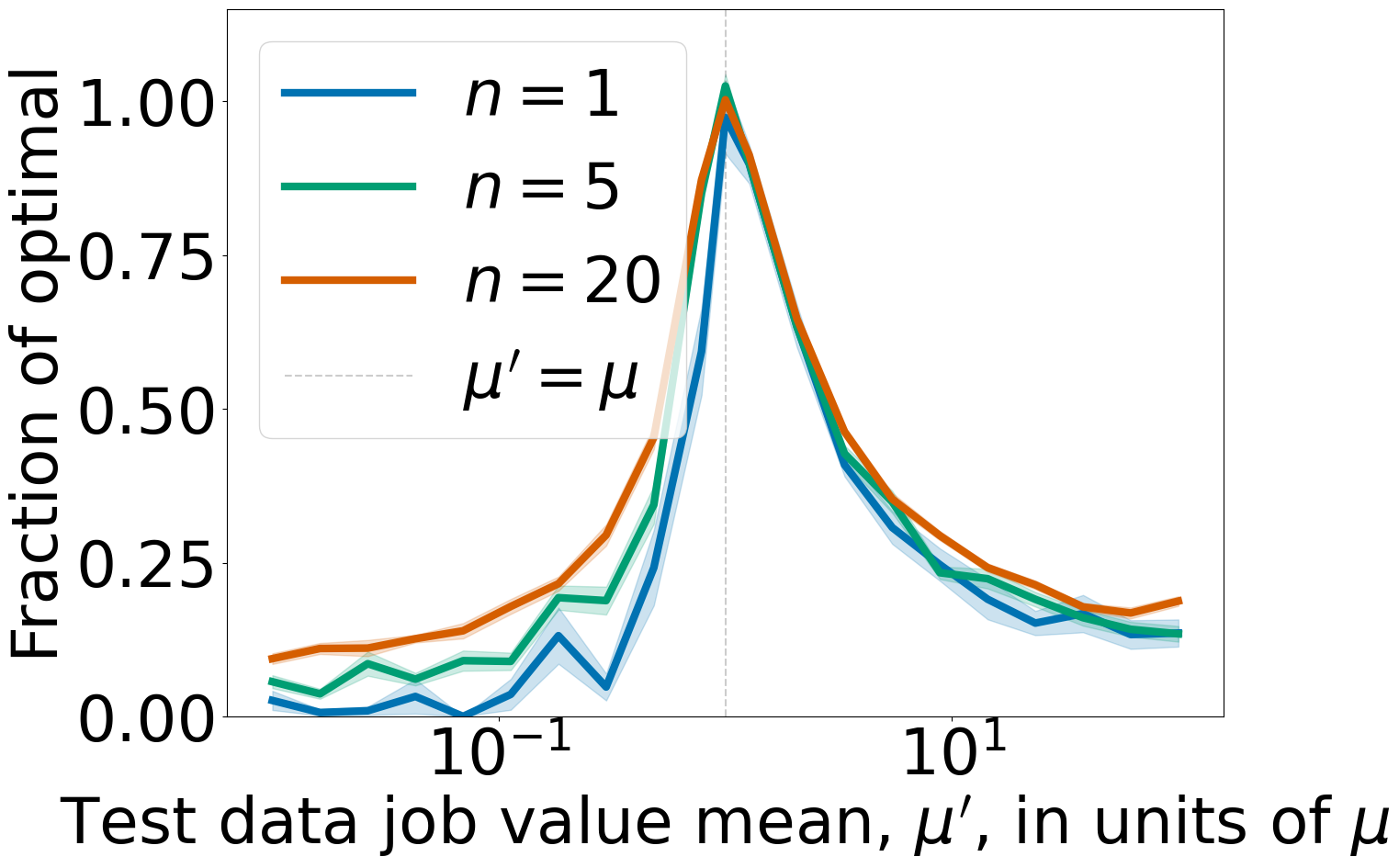}
    \caption{
    Robustness experiments for NPSA.
    Expected reward is evaluated on the processes obtained by independently varying $\lambda'$ and $\mu'$ at test time.
    A $y$-value of 1 corresponds to the best possible expected reward.
    The curvature of the plots at $x = 10^0 = 1$ shows how robust the algorithm is with respect to changes in arrival intensity (left) and value mean (right), with small curvature indicating robustness and large curvature showing the opposite.
    }
    \label{fig:expt:varying_mu_lambda}
\end{figure}

\paragraph{Convergence to Optimality.}\label{sec:expt:convergence_to_optimality}
%
%
We require a job value distribution $F$ and arrival intensity $\lambda(t)$ such that we can derive the optimal reward \emph{exactly}. 
We fit $\widetilde{\phi}$ and $\widetilde{\lambda}$ using $M$ simulated realisations of the job arrival process.
The reward observed from using NPSA-derived thresholds on further simulations is then compared to the known optimal reward as $M$ grows.

Part of the motivation for the development of NPSA stems from the intractability of exactly solving the system of ODEs necessitated by Theorem~\ref{thm:albright} for the optimal critical curves.
This strictly limits the $F$ and $\lambda(t)$ that we can use for this experiment.
Thus, we restrict the job arrival process to be homogeneous, that is, $\lambda(t) = \lambda$ for all $t \in [0, T]$.
We consider two job-value distributions, \emph{i.} exponential, that is,
\begin{equation*}
    F(z) = 1 - e^{- \frac{z}{\mu}}, \ \  \phi(z) = \mu e^{- \frac{z}{\mu}},   
\end{equation*}
where $\mu$ is the mean job value; and \emph{ii.} Lomax, that is,
\begin{equation*}\textstyle
    F(z) = 1 - (1 + \frac{z}{\xi})^{-\alpha}, \ \  \phi(z) = \frac{\xi^{\alpha} z + \xi^{\alpha + 1}}{{\left(\alpha - 1\right)} {\left(\xi + z\right)}^{\alpha}},
\end{equation*}
where $\alpha > 0$ is the shape parameter and $\xi > 0$ is the scale.
The exponential distribution is the ``simplest'' distribution in the maximum entropy sense for a nonnegative r.v. with known mean.
Lomax-distributed r.v.s are related to exponential r.v.s by exponentiation and a shift and are heavy-tailed.

Using the SageMath~\cite{sagemath} interface to Maxima~\cite{maxima}, we are able to symbolically solve for the optimal thresholds (using Theorem~\ref{thm:albright}) when $n \leq 20$ for exponentially distributed job values and $n=1$ for Lomax-distributed job values.
The optimal reward $r^\star$ is computed using the identity $r^\star = \sum^{n}_{k=1}y_k(0)$ from~\eqref{eq:expected_reward_thresholds}.
We then simulate the job arrival process for $M \in \{1, \ldots, 100\}$ independent realisations.
For each $M$, NPSA critical curves are derived using the $M$ realisations.
Then, using the same data-generating process $M' = 50$ independent realisations are played out, recording the cumulative reward obtained.
The empirical mean reward over the $M'$ simulations is computed along with its standard error and is normalised relative to $r^\star$, for each $M$.

The result is plotted in Figure~\ref{fig:varying_M_expt} for both exponentially and Lomax-distributed jobs.
We observe that in both cases convergence is rapid in $M$.
Convergence is quicker in the exponential case, which we attribute to the lighter tails than in the Lomax case, where outsize job values are more often observed that may skew the empirical estimation of $\phi$.
In the exponential case we observe that convergence is quicker as $n$ increases, which we attribute to noise from individual workers' rewards being washed out by their summation.
We further note that we have observed these qualitative features to be robust to variation of the experimental parameters.

\paragraph{Data Distribution Shift.}\label{sec:expt:data_distribution_shift}
%
%

In this experiment, jobs arrive over time horizon $T=2\pi$ according to a homogeneous Poisson process with fixed intensity $\lambda = 500$ and have values that are exponentially distributed with mean $\mu=200$.
We simulate $M=30$ realisations of the job arrival process and derive critical curves via NPSA.
We then compute modifiers $\delta_j$ for $j\in \{1, \ldots, 20\}$, where the $\delta_j$ are logarithmically spaced in the interval $[10^{-2}, 10^{2}]$.
The modifiers $\delta_j$ give rise to $\lambda'_j = \delta_j \cdot \lambda$ and $\mu'_j = \delta_j \cdot \mu$.
We fix a $j \in \{1, \ldots, 20\}$.
Holding $\mu$ (resp. $\lambda$) constant, we then generate $M' = 20$ realisations of the job arrival process with arrival rate $\lambda'_j$ (resp. mean job value $\mu'_j$) during which we accept jobs according to the thresholds derived by NPSA for $\mu$, $\lambda$.
The mean and standard error of the reward over the $M'$ realisations is recorded and normalised by the optimal reward for the true data generating process at test time, ${r'}^\star$.

The result is shown in Figure~\ref{fig:expt:varying_mu_lambda}.
Note first that the reward is very robust to variations in arrival rate. 
Indeed, using thresholds that have been derived for an arrival process where the rate differs by an order of magnitude (either an increase or decrease) incurs a relatively small penalty in reward (up to 60\%), especially when the jobs at test time arrive more frequently than during training time.
The reward is less robust with respect to variations in the mean of the job value, wherein a difference by an order of magnitude corresponds to $\approx 80\%$ loss of reward when $n=20$.
Nevertheless, for more modest deviations from the true $\mu$ value the reward is robust.


\paragraph{Evaluation on Public Fraud Data.}\label{sec:expt:public_fraud_data}
%
%
%
%
%
We augment the \textsf{SeqAlloc-NP} problem setup in Section~\ref{sec:setup} with the following.
Each job (transaction) is endowed with a feature $x \in \mathcal{X}$ and a true class label $y \in \{0, 1\}$.
The decision-maker has access to $x$ when a job arrives, but not the true label $y$. 
They also have access to a \emph{discriminator}, $D: \mathcal{X} \to [0, 1]$ that represents a subjective assessment of probability of a job with side information $x$ being a member of the positive class, $\mathbb{P}[y=1 \,\vert\, x]$.
The \emph{adjusted value} of a job $V(x, v)$, where $v$ is the job value, is given by the \emph{expected utility},  $V(x, v) = D(x) \cdot v$, where we stipulate that a job being a member of the positive class yields utility $v$, being a member of the negative class yields zero utility and the decision-maker is risk-neutral.
The decision-maker now seeks to maximise total expected utility.

\begin{table}
\setlength\tabcolsep{3pt} 
\scriptsize
\centering
\begin{tabularx}{\columnwidth}{lcccccc} 
\toprule
Dataset  & $M_{\text{train}}$ & $M_{\text{test}}$ & 
$N_{\text{daily}}^{\text{tot}}$ & 
$N_{\text{daily}}^{\text{fraud}}$ &  $v_{\text{daily}}^{\text{fraud}}$ & \texttt{clf} $F_1$-score \\
\midrule
\texttt{cc-fraud} & 2 & 1 & 94,935 & 122 & 17,403 & 0.9987  \\
\addlinespace
\texttt{ieee-fraud} & 114 & 69 & 2,853\,$\pm$\,54 & 103\,$\pm$\,4 & 16,077\,$\pm$\,738 & 0.8821 \\
\bottomrule
\end{tabularx}
\caption{Dataset properties for Figure~\ref{fig:expt:fraud_datasets} experiments.
Each dataset has $M_{\text{train}}$ realisations of training data and $M_{\text{test}}$ realisations for testing.
There are $N_{\text{daily}}^{\text{tot}}$ transactions per day in the test data, out of which $N_{\text{daily}}^{\text{fraud}}$ are fraudulent, with a total monetary value of $v_{\text{daily}}^{\text{fraud}}$.
We indicate the $F_1$-score of the \texttt{clf} classifier on the training set.
}
\label{tab:datasets}
\end{table}

\begin{figure}
    \centering
    \includegraphics[width=0.475\columnwidth]{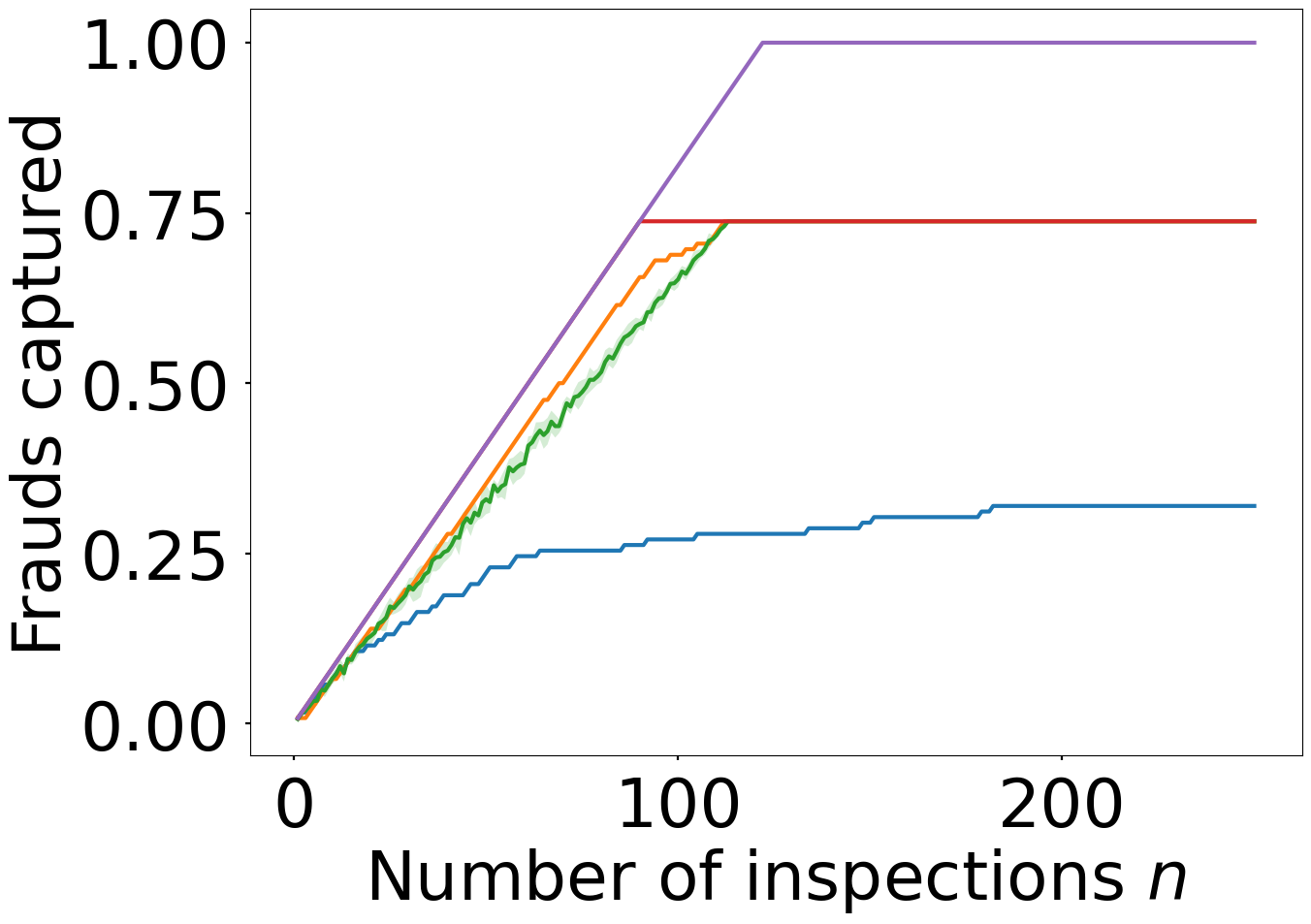}
    \includegraphics[width=0.475\columnwidth]{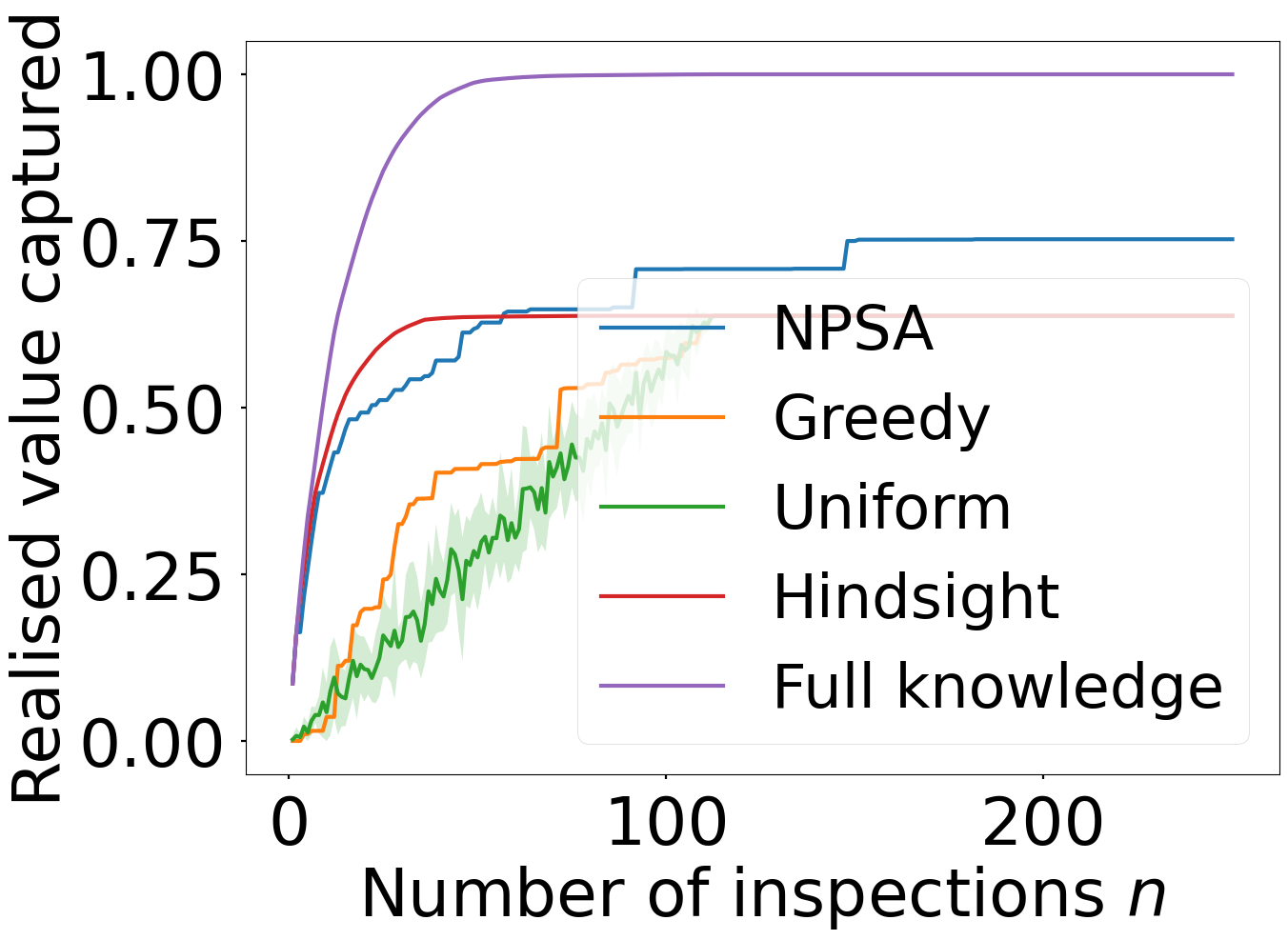}
    \includegraphics[width=0.475\columnwidth]{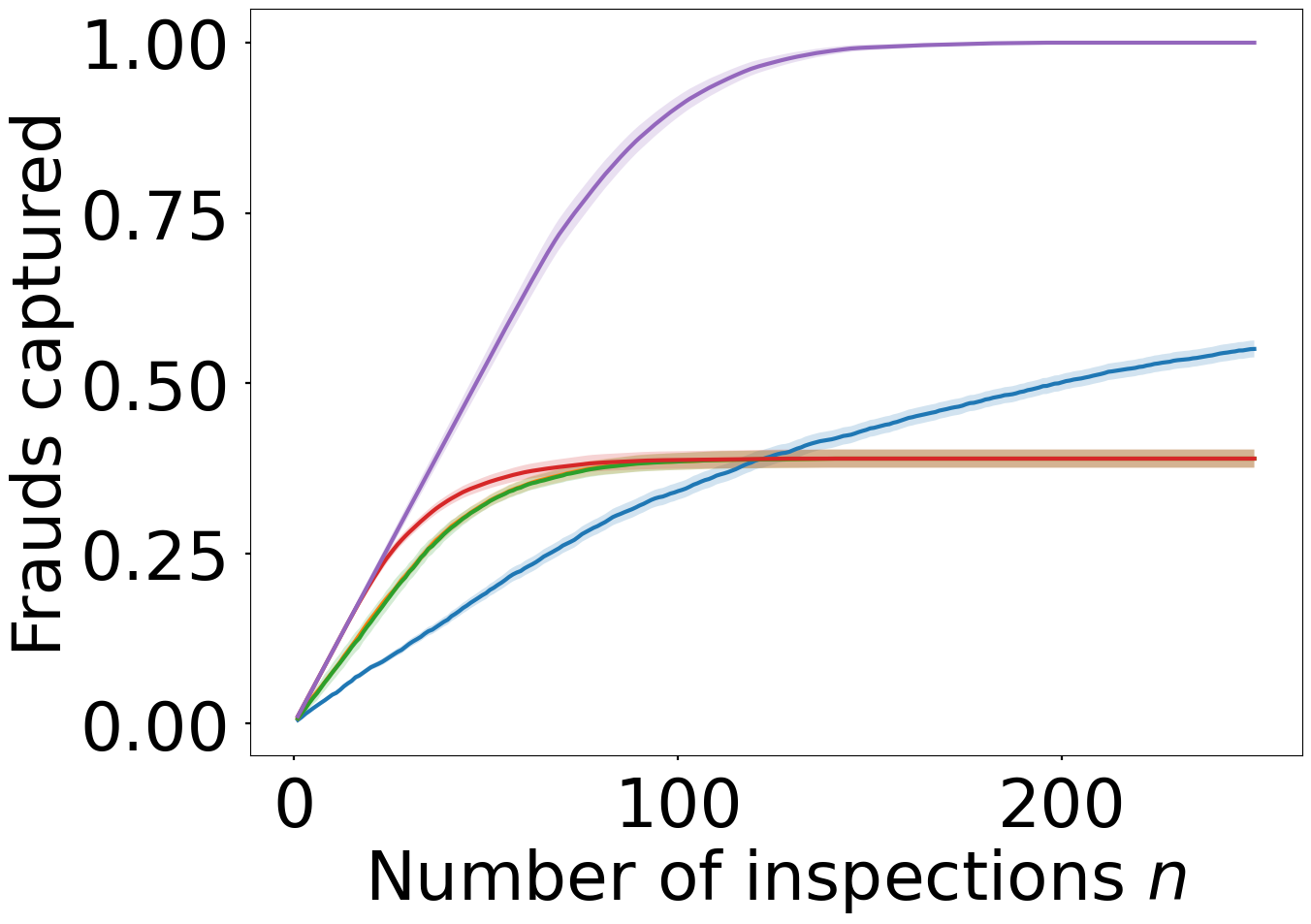}
    \includegraphics[width=0.475\columnwidth]{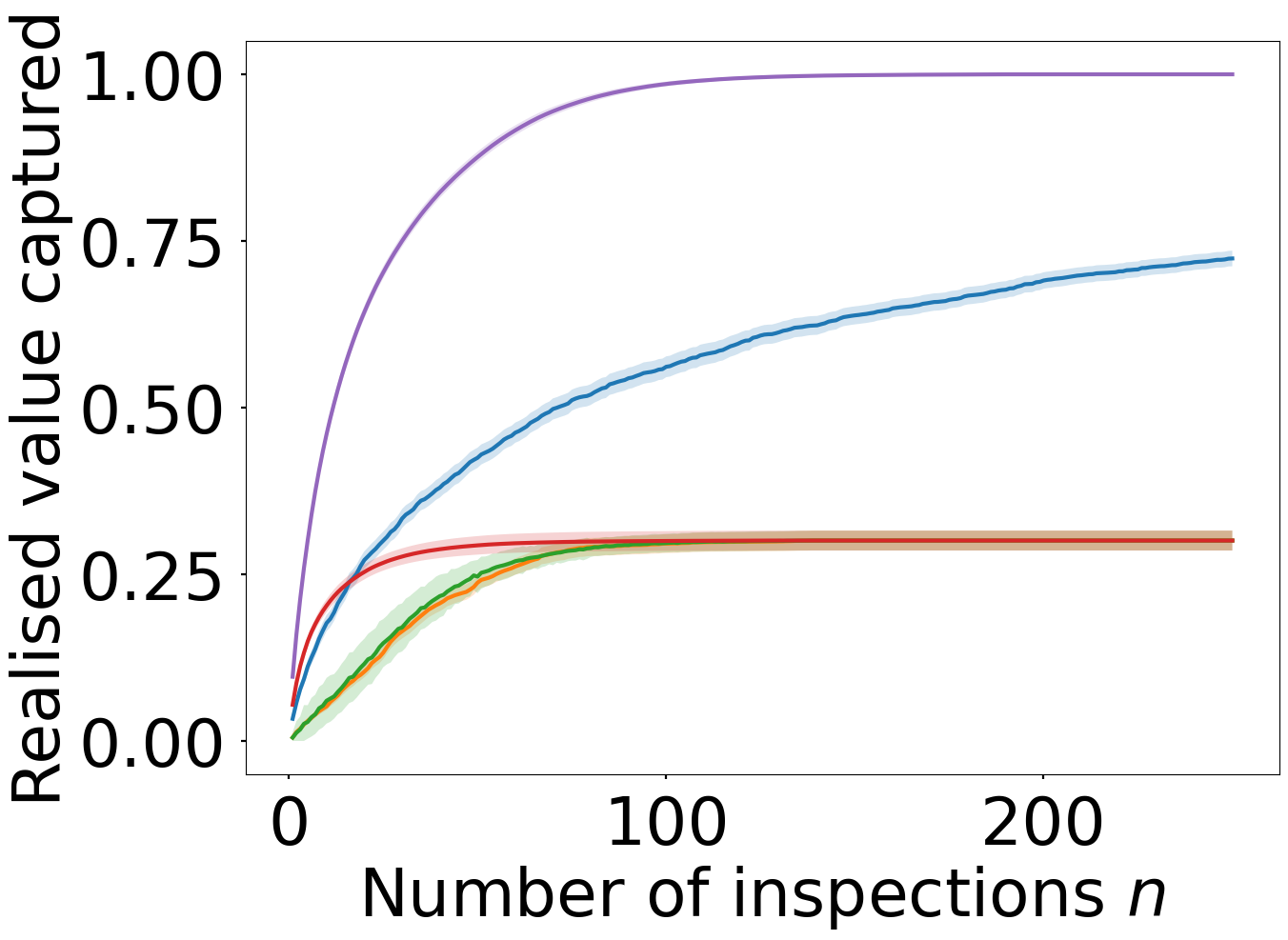}
    \caption{Fraud detection results for \texttt{cc-fraud} (top) and \texttt{ieee-fraud} (bottom) datasets.
    The left plots show the fraction of daily fraudulent transactions captured and the right show the fraction of fraudulent monetary value captured. 
    }
    \label{fig:expt:fraud_datasets}
\end{figure}

The reference time frame $T$ is set to one day and the individual realisations are split into $M_{\text{test}}$ test and $M_{\text{train}}$ training realisations, ensuring all test realisations occur chronologically after all training realisations.
A classifier \texttt{clf} is trained on the transactions in the training set using the $F_1$-score as a loss function.
This yields the discriminator $D(\, \cdot \,) \equiv \texttt{clf.predict\_proba(} \cdot \texttt{)}$ where \texttt{clf} has a scikit-learn~\cite{scikit-learn}-type interface.
For each transaction with side information $x$ and value $v$, we compute the adjusted value $V(x, v)$.
The mean-shortage function $\widetilde{\phi}$ for the adjusted job value distribution is learned on this data via the scheme described in Section~\ref{sec:phi_via_eccdf}.
Given $\widetilde{\lambda}(t)$ and $\widetilde{\phi}$, we derive critical curves via NPSA for $n \in \{1, \ldots, 250\}$, which are replayed on the $M_{\text{test}}$ test realisations.
Full details of the \texttt{clf} training and data preparation procedure are given in the Supplementary Material.


We are interested in two quantities: \emph{i.} the total monetary value of inspected transactions that are truly fraudulent, which we call \emph{realised value} and \emph{ii.} how many are truly fraudulent, or \emph{captured frauds}.
We compare these quantities obtained from the NPSA algorithm with those obtained from a number of baselines, in order of increasing capability:
\emph{i. Greedy.} Choose the first $n$ transactions \texttt{clf} marks as having positive class;
\emph{ii. Uniform.} From all the transactions \texttt{clf} marks as positive class, choose $n$ transactions uniformly at random;
\emph{iii. Hindsight.} From all the transactions \texttt{clf} marks as positive class, choose the $n$ transactions with highest monetary value;
\emph{iv. Full knowledge.} From all the transactions with $y=1$, choose the $n$ transactions with highest monetary value.
Note that \emph{iv.} is included to serve as an absolute upper-bound on performance.
We use two public fraud detection datasets, which we denote \texttt{cc-fraud}~\cite{creditcardfraud} and \texttt{ieee-fraud}~\cite{ieee-fraud-detection}.
The relevant dataset properties are given in Table~\ref{tab:datasets}.

The results are shown in Figure~\ref{fig:expt:fraud_datasets}.
First observe that NPSA shows favourable results even when trained on two realisations ($M_{\text{train}} = 2$ for the \texttt{cc-fraud} dataset), outperforming even the \emph{Hindsight} baseline for $n \geq 60$ in terms of captured realised value.
On the \texttt{ieee-fraud} dataset with $M_{\text{train}} = 114$, NPSA is outperformed only by \emph{Full Knowledge} after $n \geq 15$.
In terms of the number of captured frauds, the intuition that NPSA is waiting to inspect only the most valuable transactions to select is validated, evidenced by the the NPSA curve in these plots lying below the baseline curves, contrasted with the high realised value.

\section{Conclusion}
In this work we introduce the \textsf{SeqAlloc-NP} problem and its efficient, provably optimal solution via the NPSA algorithm.
Given $M$ independent realisations of a job arrival process, we are able to optimally select the $n$ most valuable jobs in real-time assuming the incoming data follows the same arrival process.
This algorithm is robust to variations in the data-generating process at test-time and has been applied to the financial fraud detection problem, when the value of each transaction is evaluated in a risk-neutral manner.

Future work will go down several paths: including investigating risk-hungry and risk-averse decision-makers; studying adversarial job arrival processes; addressing the effect of jobs taking up a finite time; and specialising to different application domains.

{
\paragraph{{Disclaimer.}}
This paper was prepared for informational purposes by
the Artificial Intelligence Research group of JPMorgan Chase \& Co. and its affiliates (``JP Morgan''),
and is not a product of the Research Department of JP Morgan.
JP Morgan makes no representation and warranty whatsoever and disclaims all liability,
for the completeness, accuracy or reliability of the information contained herein.
This document is not intended as investment research or investment advice, or a recommendation,
offer or solicitation for the purchase or sale of any security, financial instrument, financial product or service,
or to be used in any way for evaluating the merits of participating in any transaction,
and shall not constitute a solicitation under any jurisdiction or to any person,
if such solicitation under such jurisdiction or to such person would be unlawful.
}

\clearpage
\bibliographystyle{named}
\bibliography{references}

\clearpage
\appendix

\begin{strip} 
\begin{center}
    {\LARGE\bf Supplementary Material for \emph{Non-parametric stochastic sequential assignment with random arrival times} }
\end{center}
\end{strip}

\section{Mean Shortage Function Estimator Pseudocode}

\begin{algorithm}
\DontPrintSemicolon
\caption{Mean Shortage Function Approximator}
\label{alg:ecdf_mean_shortage}
\SetKwInOut{Input}{Input}
\Input{$y \in [0, \infty)$}
\KwData{Reward samples $0 < x_1 < \ldots < x_N$}
\SetKwInOut{Output}{Output}
\Output{$\widetilde{\phi}_{N}(y) \approx \phi(y)$}
\BlankLine
\SetKwProg{buildcache}{build\_cache}{}{end}
\buildcache(\tcp*[f]{run once}){$(x_1, \ldots, x_N)$}{
    $\phi_{N} \leftarrow 0$\;
    \For{$i$ \KwSty{in} $(N - 1, \ldots, 1)$}{
        $\phi_i \leftarrow \phi_{i+1} + (x_{i+1} - x_i) \cdot \frac{N - i}{N}$
    }
}
\Begin(\tcp*[f]{at eval time}){
    \lIf{$y \geq x_N$}{
    \KwRet{$0$}}
    \lElseIf{$y < x_1$}{
    \KwRet{$(x_1 - y) \phi_{1}$}
    }
    \Else{
    Find $\ell$ such that $x_{\ell} \leq y \leq x_{\ell + 1}$ via binary search\;
    $\widetilde{\phi} \leftarrow \phi_{\ell + 1}$ \;
    \KwRet{$\widetilde{\phi} + (x_{\ell + 1} - y) \phi_{\ell}$}
    }
}
\end{algorithm}

\section{Technical Proofs}

\subsection{Proof of Lemma~\ref{lem:mean_shortage_integral}}
\begin{proof}
We use a similar technique to~\citeSM{SE_exp_value_cdf}.
Write
$\int^\infty_y (1 - F(x)) \dd x = \int^\infty_y \mathbb{P}(X \geq x) \dd x = \int^\infty_y \int^\infty_x f(t) \dd t \dd x$.
Switching the order of integration yields $\int^\infty_y \int^t_y f(t) \dd x \dd t = \int^\infty_y \qty[  x f(t)]^t_y \dd t = \int^\infty_y (t - y) f(t) \dd t$.
Recognising $t$ as a dummy variable and renaming to $x$ produces the result.
\end{proof}

\subsection{Proof of Theorem~\ref{thm:ecdf_convergence}}
\begin{proof}
Denote the cdf of $X$ by $F$ and the empirical cdf computed from $N$ independent samples of $X$ by $F_N(x)$.
It follows from the specification of Algorithm~\ref{alg:ecdf_mean_shortage} that $\widetilde{\phi}_{N}(y) = \int^\infty_y (1 - F_N(x)) \dd x$ as the integral of a piecewise constant function is the sum of the areas of the corresponding rectangles and $F_N$ is a piecewise constant function with domain $[0, \infty)$.

Now suppose that $\sup_{x \in [0, \infty)} \abs{F(x) - F_N(x)} < \gamma$ for some $\gamma > 0$. Then, for any $y \in [0, \infty)$,
\begin{talign}
    &\abs*{\widetilde{\phi}_{N}(y) - \phi(y)} \nonumber \\
    &\quad= \abs{ \int^\infty_y (1 - F(x)) \dd x - \int^\infty_y (1 - F_N(x)) \dd x } \nonumber \\
    &\quad= \Big\vert \qty( \phi(0) - \int^y_0 (1 - F(x)) \dd x ) \nonumber \\
    &\quad\hspace{2.3cm}- \qty( \widetilde{\phi}_{N}(0) - \int^y_0 (1 - F_N(x)) \dd x ) \Big\vert \nonumber \\
    &\quad\leq \abs*{\phi(0) - \widetilde{\phi}_{N}(0)} + \abs{\int^y_0 (F_N(x) - F(x)) \dd x} \nonumber \\
    &\quad\leq \abs*{\mu - \widetilde{\mu}^{(N)}} + \int^y_0 \abs{F_N(x) - F(x)} \dd x \nonumber \\
    &\quad\leq \abs*{\mu - \widetilde{\mu}^{(N)}} + \int^y_0 \gamma \dd x \nonumber \\
    &\quad= \abs*{\mu - \widetilde{\mu}^{(N)}} + \gamma y, \label{eq:ecdf_consistency_ineq}
\end{talign}
where $\widetilde{\mu}^{(N)} := \frac{1}{N}\sum_{i=1}^N x_i$ is the sample mean of $X$.
We will show that $\lim_{N \to \infty} \mathbb{P} \qty[ \abs*{\widetilde{\phi}_{N}(y') - \phi(y')} > \epsilon] = 0$ for all $y\geq 0$, from which the result follows.
We distinguish two cases: \emph{i.} $y=0$ and \emph{ii.} $y > 0$.

\emph{Case i.} If $y = 0$ then $\abs*{\widetilde{\phi}_{N}(y) - \phi(y)} \leq \abs*{\mu - \widetilde{\mu}^{(N)}}$ by~\eqref{eq:ecdf_consistency_ineq}, which converges almost surely to zero as $N \to \infty$ by the strong law of large numbers.
In this case the result follows immediately, since almost sure convergence implies convergence in probability.

\emph{Case ii.} Now suppose $y > 0$, fix an arbitrary $\epsilon > 0$ and assume that $\abs*{\widetilde{\phi}_{N}(y) - \phi(y)} > \epsilon$.
Moreover, introduce the variable parameter $\delta > 0$.
Either $\abs{F(y) - F_N(y)} \geq \delta$ or $y \in B_\delta$, where we define $B_\delta := \left\{x \in [0, \infty) \middle\vert \abs{F(x) - F_N(x)} < \delta,\ \abs*{\widetilde{\phi}_{N}(x) - \phi(x)} > \epsilon \right\}$.
Extending to many realisations of the $N$ samples of $X$,
\begin{multline}\label{eq:ecdf_consistency_prob_sum}
    \mathbb{P}\left[\abs*{\widetilde{\phi}_{N}(y) - \phi(y)} > \epsilon \right] \leq \mathbb{P}\left[\abs{F(y) - F_N(y)} \geq \delta \right] \\+ \mathbb{P}\left[y \in B_\delta \right].
\end{multline}
From the strong law of large numbers, for any $\xi > 0$ there exists a $N_\xi$ such that for all $N > N_\xi$, $\abs*{\mu - \widetilde{\mu}^{(N)}} < \xi$.
We now assume that $N > N_\epsilon$, such that $\epsilon > \abs*{\mu - \widetilde{\mu}^{(N)}}$.
Setting $\delta \leq (\epsilon - \abs*{\mu - \widetilde{\mu}^{(N)}})\cdot y^{-1}$ gives us $B_\delta = \emptyset$ from~\eqref{eq:ecdf_consistency_ineq} and so $\mathbb{P}\left[y \in B_\delta \right] = 0$.
Substituting into~\eqref{eq:ecdf_consistency_prob_sum} yields
\begin{talign}
    \label{eq:ecdf_consistency_preDKW}
    &\mathbb{P}\left[ \abs*{\widetilde{\phi}_{N}(y) - \phi(y)} > \epsilon \right] \nonumber \\
    &\quad\leq \mathbb{P}\left[\abs{F(y) - F_N(y)} \geq \frac{\epsilon - \abs*{\mu - \widetilde{\mu}^{(N)}}}{y} \right] \nonumber \\ 
    &\quad\leq \mathbb{P}\left[\sup_{z \in [0, \infty)} \abs{F(z) - F_N(z)} \geq \frac{\epsilon - \abs*{\mu - \widetilde{\mu}^{(N)}}}{y} \right],
\end{talign}
where the second inequality follows from the fact that $\abs{F(x) - F_N(x)} \geq \xi \Rightarrow \sup_{z \in [0, \infty)} \abs{F(z) - F_N(z)} \geq \xi$ for all $x \in [0, \infty)$, $\xi > 0$.
Employing the Dvoretzky–Kiefer–Wolfowitz inequality~\cite{dvoretzky1956,massart1990} in~\eqref{eq:ecdf_consistency_preDKW} yields
\begin{equation}\label{eq:ecdf_consistency_converge}
    \mathbb{P}\left[\abs*{\widetilde{\phi}_{N}(y) - \phi(y)} > \epsilon \right] 
    \leq  2\exp( - \textstyle\frac{2N\qty(\epsilon - \abs*{\mu - \widetilde{\mu}^{(N)}})^2}{y^2}).
\end{equation}
Now choose $\zeta$ such that $0 < \zeta < \epsilon$.
We have from the strong law of large numbers that there exists $N_\zeta > N_\epsilon$ such that $\abs*{\mu - \widetilde{\mu}^{(N)}} < \zeta < \epsilon$ for all $N > N_\zeta$.
Therefore, when $N > N_\zeta$ we can bound~\eqref{eq:ecdf_consistency_converge} by
\begin{equation*}
    \mathbb{P}\left[\abs*{\widetilde{\phi}_{N}(y) - \phi(y)} > \epsilon \right] 
    \leq  2\exp( - \textstyle\frac{2N\qty(\epsilon - \zeta)^2}{y^2})
\end{equation*}
Finally passing to the limit as $N \to \infty$, we have that
$$\lim_{N \to \infty} \mathbb{P} \qty[ \abs{\widetilde{\phi}_{N}(y) - \phi(y)} > \epsilon] = 0$$
from $(\epsilon - \zeta)$ and $y$ being fixed and strictly positive. 
The result follows, since $y > 0$ is arbitrary.
\end{proof}

\subsection{Proof of Lemma~\ref{lem:y_prime_epsilon_approx}}
\begin{proof}
We have from the definition of an $\epsilon$-approximator and from Theorem~\ref{thm:albright} that $y'_k$ $\epsilon$-approximates $y_k$ (we omit the explicit reference to $t$ for brevity) for all $k \in \{1, \ldots, n\}$  when
\begin{equation}\label{eq:eps_approx_1}
    \abs{\lambda'(t)\qty( \phi'(y'_{k+1}) - \phi'(y'_k) ) - \lambda(t)\qty( \phi(y'_{k+1}) - \phi(y'_k) ) } < \epsilon
\end{equation}
for all $t \in [0, T]$.
We shall proceed by upper-bounding the left-hand side of~\eqref{eq:eps_approx_1}.

Recall that $y'_{k+1} \leq y'_k$ from Theorem~\ref{thm:albright} and that $\phi'$ is a nonincreasing function from its definition.
Thus, $\phi'(y'_{k+1}) - \phi'(y'_k) \geq 0$.
The left hand side of~\eqref{eq:eps_approx_1} is equal to
\begin{align}
    & \abs{\lambda(t)\qty( \phi(y'_{k+1}) - \phi(y'_k) ) - \lambda'(t)\qty( \phi'(y'_{k+1}) - \phi'(y'_k) ) }  &\nonumber \\
    &\quad\leq \big\vert \lambda(t)\qty( \phi(y'_{k+1}) - \phi(y'_k) )       \nonumber \\
    &\hphantom{\qty( \phi(y'_{k+1}) - \phi(y'_k) )}    - (1 - \delta_\lambda) \lambda(t) \qty( \phi'(y'_{k+1}) - \phi'(y'_k) ) \big\vert   \nonumber \\
    &\quad= \lambda(t) \big\vert\qty( \phi(y'_{k+1}) - \phi(y'_k) ) \nonumber \\
    &\hphantom{\qty( \phi(y'_{k+1}) - \phi(y'_k) )} - (1 - \delta_\lambda) \qty( \phi'(y'_{k+1}) - \phi'(y'_k) ) \big\vert.  \label{eq:eps_approx_2}
\end{align}
Furthermore, $-\phi'(y'_{k+1}) \leq - (1 - \delta_\phi) \phi(y'_{k+1})$ and $\phi'(y'_k) \leq (1 + \delta_\phi) \phi(y'_k)$, so
\begin{align*}
    - (\phi'(y'_{k+1}) - \phi'(y'_k) ) &\leq - (1 - \delta_\phi) \phi(y'_{k+1})\\
    &\hphantom{\phi(y'_{k+1}) +l}+ (1 + \delta_\phi) \phi(y'_k) \\
    &\leq - \qty(\phi(y'_{k+1}) - \phi(y'_k) ) \\
    &\hphantom{\phi(y'_{k+1}) +l}+ \delta_\phi \qty(\phi(y'_{k+1}) + \phi(y'_{k})).
\end{align*}
Substituting into~\eqref{eq:eps_approx_2} we have
\begin{align*}
    &\lambda (t) \abs{ \qty( \phi(y'_{k+1}) - \phi(y'_k) ) - (1 - \delta_\lambda ) \qty( \phi'(y'_{k+1}) - \phi'(y'_k) ) } \\
    &\quad\leq  \lambda (t) \big\vert \qty( \phi(y'_{k+1}) - \phi(y'_k) ) + (1 - \delta_\lambda )  \times \\ 
    &\quad \hphantom{\lambda(t)hl}\hspace{0.6mm} \qty[- \qty(\phi(y'_{k+1}) - \phi(y'_k) ) + \delta_\phi \qty(\phi(y'_{k+1}) + \phi(y'_{k}))] \big\vert & \\
    &\quad= \lambda (t) \big\vert \delta_\phi (1 - \delta_\lambda) \qty(\phi(y'_{k+1}) + \phi(y'_{k})) \\
    &\quad \hphantom{hhhhhhhhhhhhhhhhhhhl|.}\hspace{1.4mm} + \delta_\lambda  \qty(\phi(y'_{k+1}) - \phi(y'_k) ) \big\vert & \\
    &\quad\leq \lambda (t) \abs{ \delta_\phi \qty(\phi(y'_{k+1}) + \phi(y'_{k})) + \delta_\lambda  \qty(\phi(y'_{k+1}) - \phi(y'_k) ) } & \\
    &\quad\leq \lambda (t) \abs{ \delta_\phi \qty(\phi(y'_{k+1}) + \phi(y'_{k})) + \delta_\lambda  \qty(\phi(y'_{k+1}) + \phi(y'_k) ) } & \\
    &\quad\leq \lambda(t) (\delta_\phi + \delta_\lambda)  \qty(\phi(y'_{k+1}) + \phi(y'_k) ) \\
    &\quad\leq 2 \mu \lambda_{\text{max}} (\delta_\phi + \delta_\lambda), 
\end{align*}
where we have used the fact that $\phi(y') \leq \mu$ for all $y' \in [0, \infty)$ for the final inequality.
\end{proof}

\subsection{Proof of Lemma~\ref{lem:threshold_dist_bound}}
\begin{proof}
    We proceed by way of Lemma~\ref{lem:eps_approx_dist}.
    Recall from Theorem~\ref{thm:albright} that $y_k(t)$ is solved via an initial value problem starting at $t=T$ and proceeding back in time until $t=0$.
    We are led to
    \begin{multline}\label{eq:thresh_opt_1}
        \abs{y_k(t) - y'_k(t)} \leq \abs{y_k(T) - y'_k(T)} e^{L_g (T - t)} + \\ \frac{\epsilon}{L_g} \qty(e^{L_g (T - t)} - 1),
    \end{multline}
    assuming $y'_k(t)$ is an $\epsilon$-approximator to $y_k(t)$ and denoting by $L_g$ a Lipschitz constant of the function
    \begin{equation*}
        g(t, x_1, x_2) := \lambda(t) \qty(\phi(x_1) - \phi(x_2)), \qq{} x_1, x_2 \in [0, \infty).
    \end{equation*}
    Recall from Theorem~\ref{thm:albright} that $y_k(T) = y'_k(T) = 0$.
    Moreover, from Lemma~\ref{lem:y_prime_epsilon_approx}, we have that $\epsilon > 2 \mu \lambda_{\text{max}} (\delta_\lambda + \delta_\phi)$.
    Substitution into~\eqref{eq:thresh_opt_1} yields
    \begin{equation}\label{eq:thresh_opt_2}
        \abs{y_k(t) - y'_k(t)} \leq \frac{2 \mu \lambda_{\text{max}} (\delta_\lambda + \delta_\phi)}{L_g} \qty(e^{L_g (T - t)} - 1)
    \end{equation}
    It remains to prove that $2\lambda_{\text{max}}$ is a Lipschitz constant for the function $g(t, x_1, x_2)$, which we shorten to $g$ for brevity.
    
    To wit, $\lambda(t) \geq 0$ for all $t \in [0, T]$, so we have that $\lambda_{\text{max}} L_{\widetilde{g}}$ is a Lipschitz constant for $g$, where
    \begin{equation*}
        \widetilde{g}(x_1, x_2) := \phi(x_1) - \phi(x_2), \qq{} x_1, x_2 \in [0, \infty)
    \end{equation*}
    and $L_{\widetilde{g}}$ is a Lipschitz constant for $\widetilde{g}$.
    The function $\widetilde{g}$ is comprised of the difference of two convex, nonincreasing functions and so $L_{\widetilde{g}}$ can be given by the sum of their Lipschitz constants, that is, $L_{\widetilde{g}} = 2 L_\phi$.
    We must find an $L_\phi$ such that $\abs{\phi(z) - \phi(z')} \leq L_\phi \abs{z - z'}$ for $z, z' \in [0, \infty)$.
    Upper-bounding the left hand side,
    \begin{align*}
        & \abs{\phi(z) - \phi(z')}  \\
        &\quad\leq \abs{\textstyle\int_{z}^\infty(1 - F(x)) \dd x - \textstyle\int_{z'}^\infty(1 - F(x)) \dd x} \\
        &\quad= \abs{\textstyle\int_{z}^{z'}(1 - F(x)) \dd x } \leq \textstyle\int_{z}^{z'}\abs{1 - F(x)} \dd x \\
        &\quad\leq \abs{\textstyle\int_{z}^{z'}\ \dd x} = \abs{z - z'}
    \end{align*}
    where the inequality in the last line follows from $0\leq F(x) \leq 1$ for all $x \in [0, \infty)$. 
    We are thus able to set $L_\phi = 1$.
    As a result, $2\lambda_{\text{max}}$ is a Lipschitz constant for $g$, and the result follows.
\end{proof}

\subsection{Proof of Lemma~\ref{lem:reward_lower_bound}}
\begin{proof}
We shall proceed via induction on $n$, lower bounding individual terms in~\eqref{eq:expected_reward_integral} to produce the result.
First, we bound
\begin{align}\label{eq:lem:reward_lower_bound_eq1}
    &\exp[- \textstyle\int^\tau_t \lambda(\sigma) \overline{F}(y'_k(\sigma)) \dd \sigma ] \nonumber \\ 
    &\quad\geq \exp[- \textstyle\int^\tau_t \lambda(\sigma) \qty(\overline{F}(y_k(\sigma))- \epsilon_{\overline{F}} ) \dd \sigma ] \nonumber \\
    &\quad= \exp[- \textstyle\int^\tau_t \lambda(\sigma) \overline{F}(y_k(\sigma)) \dd \sigma ]\exp[- \epsilon_{\overline{F}}\textstyle\int^\tau_t \lambda(\sigma)\dd \sigma ] \nonumber \\
    &\quad\geq \exp[- \textstyle\int^\tau_t \lambda(\sigma) \overline{F}(y_k(\sigma)) \dd \sigma ]\exp[- \epsilon_{\overline{F}}\textstyle\int^T_0 \lambda(\sigma)\dd \sigma ] \nonumber \\
    &\quad= \exp[- \textstyle\int^\tau_t \lambda(\sigma) \overline{F}(y_k(\sigma)) \dd \sigma ]\exp[-\epsilon_{\overline{F}} \overline{\lambda} T],
\end{align}
where we have used $\abs{\overline{F}(y'_k(\sigma)) - \overline{F}(y_k(\sigma))} \leq \epsilon_{\overline{F}}$ in the first inequality and monotonicity of $\exp(\,\cdot\,)$ in the penultimate line, along with the fact that $\lambda(t)\geq 0$.

Now we assume $n = 1$.
We have from~\eqref{eq:expected_reward_integral} and the convention that $E_0(t) = 0$ for all $t \in [0, T]$,
\begin{talign}
    &E_1(t ; y'_1) \nonumber \\
    &\quad= \int_t^T \lambda(\tau) H(y'_1(\tau)) \exp[- \textstyle\int^\tau_t \lambda(\sigma) \overline{F}(y'_1(\sigma)) \dd \sigma ] \dd \tau \nonumber \\
    &\quad\geq e^{-(\delta_H + \overline{\lambda}\epsilon_{\overline{F}} T)} \nonumber \\ 
    &\quad\hspace{0.5cm}\times\int_t^T \lambda(\tau) H(y_1(\tau)) \exp[- \textstyle\int^\tau_t \lambda(\sigma) \overline{F}(y_1(\sigma)) \dd \sigma ] \dd \tau \nonumber \\ 
    &\quad= e^{-(\delta_H + \overline{\lambda}\epsilon_{\overline{F}} T)} \cdot E_1(t ; y_1) \nonumber \\
    &\quad\geq e^{-(\delta + \overline{\lambda}\epsilon_{\overline{F}} T)} \cdot E_1(t ; y_1),
\end{talign}
where for the first inequality we have employed~\eqref{eq:lem:reward_lower_bound_eq1}, the assumption that $H(y'_1(\sigma)) \geq e^{-\delta_H}H(y_1(\sigma))$ and recalled that $\delta = \max \{\delta_{\overline{F}}, \delta_H\}$.
This proves the result for $n=1$.

We now take the inductive step and assume that $2 \leq k \leq n$.
For the first two terms in the integrand of~\eqref{eq:expected_reward_integral}
\begin{align}\label{eq:lem:reward_lower_bound_eq2}
    &H(y'_k(\tau)) +  \overline{F}(y'_k(\sigma))\cdot E_{k-1}(\tau ; y'_{k-1}, \ldots, y'_1) \nonumber \\
    &\quad\geq e^{-\delta_H} H(y_k(\tau)) + \nonumber\\
    &\quad\hspace{1.5cm} e^{-\delta_{\overline{F}}} \overline{F}(y_k(\sigma))\cdot E_{k-1}(\tau ; y'_{k-1}, \ldots, y'_1) \nonumber\\
    &\quad\geq e^{-\delta}\Big[ H(y_k(\tau)) + \nonumber \\
    &\quad\hspace{1.5cm} \overline{F}(y_k(\sigma))\cdot E_{k-1}(\tau ; y'_{k-1}, \ldots, y'_1)\Big] \nonumber\\
    &\quad\geq e^{-\delta} \Big[ H(y_k(\tau)) + \overline{F}(y_k(\sigma)) e^{-(k-1)(\delta + \overline{\lambda}\epsilon_{\overline{F}}T)}  \\
    &\quad\hspace{3.5cm} \times E_{k-1}(\tau ; y_{k-1}, \ldots, y_1) \Big]\nonumber\\
    &\quad\geq e^{-\delta} e^{-(k-1)(\delta + \overline{\lambda}\epsilon_{\overline{F}}T)}  \Big[ H(y_k(\tau)) + \overline{F}(y_k(\sigma))  \nonumber \\
    &\quad\hspace{3.5cm} \times E_{k-1}(\tau ; y_{k-1}, \ldots, y_1) \Big]\nonumber\\
    &\quad= e^{-k\delta} e^{-(k-1)(\overline{\lambda}\epsilon_{\overline{F}}T)}  \Big[ H(y_k(\tau)) + \overline{F}(y_k(\sigma)) \nonumber \\
    &\quad\hspace{3.5cm} \times E_{k-1}(\tau ; y_{k-1}, \ldots, y_1) \Big],\nonumber
\end{align}
where we have used the inductive hypothesis in the third inequality.
Substituting~\eqref{eq:lem:reward_lower_bound_eq1}~and~\eqref{eq:lem:reward_lower_bound_eq2} into~\eqref{eq:expected_reward_integral} we have
\begin{align}
    &E_k(t ; y'_k, \ldots, y'_1) \geq \nonumber \\
    &\quad e^{-\overline{\lambda}\epsilon_{\overline{F}} T} e^{-k\delta} e^{-(k-1)(\overline{\lambda}\epsilon_{\overline{F}}T)} \nonumber \\
    &\quad \times\int_t^T \bigg[ H(y_k(\tau)) +  \overline{F}(y_k(\tau))\cdot E_{k-1}(\tau ; y_{k-1}, \ldots, y_1) \bigg] \nonumber\\
    &\quad\hphantom{=\int_t^T} \times \bigg[ \lambda(\tau)\exp[- \textstyle\int^\tau_t \lambda(\sigma) \overline{F}(y_k(\sigma)) \dd \sigma ]\bigg] \dd \tau, 
\end{align}
which, on comparison with~\eqref{eq:expected_reward_integral} yields the result for arbitrary $k\leq n$.
\end{proof}

\subsection{Some Auxiliary Results for Theorem~\ref{thm:optimal_convergence}}

\begin{lemma}\label{lem:lipschitz_prod}
Let the function $g: D \to [\beta, \infty)$ be on domain $D \subseteq \mathbb{R}$, where $\beta > 0$, and let $x, y \in D$.
Furthermore, suppose $g$ is $L$-Lipschitz.
Then
\begin{talign*}\textstyle
    \abs{\ln(g(x)) - \ln(g(y))} \leq \frac{L}{\beta}\abs{x - y}
\end{talign*}
\end{lemma}
\begin{proof}
Observe that 
\begin{equation}\label{eq:lem_lipschitz_prod_eq_1}
    \abs{\ln(g(x)) - \ln(g(y))} \leq \textstyle\frac{1}{\beta} \abs{g(x) - g(y)},
\end{equation}
which follows from the derivative of $\ln(z)$, $1/z$, being upper-bounded by $1/\beta$ on $[g(y), \infty)$ and $\ln(z)$ being monotone decreasing on this domain.
The absolute value in the right-hand side of~\eqref{eq:lem_lipschitz_prod_eq_1} is upper-bounded by $L\abs{x-y}$, from which the result follows.
\end{proof}

\begin{claim}\label{claim:HFlowerbound}
    For all $t \in [0, T]$ and $k \in \{1, \ldots, n\}$,
    \begin{enumerate}
        \item $H(y_k(t)), H(y'_k(t)) \geq \beta_H$;
        \item $\overline{F}(y_k(t)), \overline{F}(y'_k(t)) \geq \beta_{\overline{F}}$;
        \item $\phi(y_k(t)), \phi(y'_k(t)) \geq \beta_\phi$,
    \end{enumerate}
    for some constant $\beta_H, \beta_{\overline{F}}, \beta_\phi > 0$.
\end{claim}
\begin{proof}
We have that $y_{k+1}(t) \leq y_k(t)$ and $y'_{k+1}(t) \leq y'_k(t)$ for all $t \in [0, T]$ from Theorem~\ref{thm:albright}.
Moreover, each critical curve is monotone decreasing.
Thus, it is sufficient to prove the statement for $y_1(0)$ and $y'_1(0)$.
We distinguish between two cases.

\emph{Case i. finite support.} 
Suppose the common support of $X$ and $X'$ has some least upper bound, $U$, such that
$U := \sup\qty{y \,\middle\vert\, F(y) < 1} = \sup\qty{y \,\middle\vert\, F'(y) <  1}$.
For the sake of contradiction let $y_1(0) \geq U$.
There is no job that can arrive that will accepted by the critical curve $y_1$ at this time, $t=0$, as $F(y_1(0)) = 1$.
Due to the finite time horizon $T < \infty$, we can increase the expected reward by reducing the critical curve $y_1(0)$ by some small $\xi > 0$, such that $F(y_1(0) - \xi) < 1$.
But $y_1$ is optimal, so we have a contradiction and therefore $y_1(0) < U$.
An identical argument holds for $y_1'(0)$.
Thus we can choose lower bounds $\beta_H = \min\qty{H(y_1(0)), H(y'_1(0))}$, $\beta_{\overline{F}} = \min\qty{\overline{F}(y_1(0)), \overline{F}(y'_1(0))}$ and  $\beta_\phi = \min\qty{\phi(y_1(0)), \phi(y'_1(0))}$ as $H$, $\overline{F}$ and $\phi$ are monotone decreasing.

\emph{Case ii. infinite support.}
Suppose the support of $X$ and $X'$ is infinite, that is, $F(y), F(y') < 1$ for all $y \geq 0$.
From Eqs~\eqref{eq:expected_reward_thresholds}~and~\eqref{eq:expected_reward_thresholds} we have
\begin{align}
    y_1(0) &= E_1(0; y_1) \nonumber \\
    &= \textstyle\int_0^T \lambda(\tau) H(y_1(\tau)) \exp[- \textstyle\int^\tau_0 \lambda(\sigma) \overline{F}(y_1(\sigma)) \dd \sigma ] \dd \tau \nonumber \\
    &\leq \mu \textstyle\int_0^T \lambda(\tau) \dd \tau \nonumber \\
    &= \mu T \overline{\lambda},
\end{align}
where the inequality on the third line follows from $H(y) \leq \mu$ for all $y \geq 0$ and $e^{-z} \leq 1$ for $z \geq 1$.
A similar argument holds for $y'_1(0)$.
Define $\mu_{\text{max}}:= \max\qty{\mu, \mu'}$.
We then choose lower bounds $\beta_H = H(\mu_{\text{max}} T \overline{\lambda})$, $\beta_{\overline{F}} = \overline{F}(\mu_{\text{max}} T \overline{\lambda})$ and $\beta_\phi = \phi(\mu_{\text{max}} T \overline{\lambda})$ since $H$, $\overline{F}$ and $\phi$ are monotone decreasing.

From their definition $H(y) > 0$, $\overline{F}(y) > 0$ and $\phi(y) > 0$ if and only if $F(y) < 1$, which is true in both cases \emph{i.} and \emph{ii.}. We have thus found the constants $\beta_{H}$, $\beta_{\overline{F}}$, $\beta_\phi$ satisfying $\min\qty{H(y_1(0)), H(y'_1(0))} \geq \beta_H > 0$, $\min\qty{\overline{F}(y_1(0)), \overline{F}(y'_1(0))} \geq \beta_{\overline{F}} > 0$ and $\min\qty{\phi(y_1(0)), \phi(y'_1(0))} \geq \beta_\phi > 0$, proving the claim.
\end{proof}
\begin{claim}\label{claim:HFLipschitz}
    Define $y_{\text{max}}:=\max\qty{y_1(0), y'_1(0)}$, $D := \big[0, y_{\text{max}}]$ and $f_{\text{max}} := \sup_{y \in D} f(y)$. 
    On the domain $D$ the functions $\overline{F}(y)$ and $H(y)$ are $L_{\overline{F}}$ and $L_H$-Lipschitz respectively, for $L_{\overline{F}} = f_{\text{max}}$ and $L_H = f_{\text{max}} y_{\text{max}}$. 
\end{claim}
\begin{proof}
For the first part, we have for  $x, y \in D$
\begin{talign}\label{eq:claim:HFLipschitz}
  \abs{\overline{F}(x) - \overline{F}(y)} &\leq \abs{1 - F(x) - 1 + F(y) } = \abs{F(y) - F(x)} \nonumber \\
  &= \abs{ \int_0^y f(z) \dd z - \int_0^x f(z) \dd z } =  \abs{\int_y^x f(z) \dd z} \nonumber \\
  &= \abs{f(c) \int_y^x \dd z} \qq{for some $c \in [x,y]$} \nonumber \\
  &\leq f_{\text{max}} \abs{x - y},
\end{talign}
where we use the mean value theorem for integrals in the penultimate line.
For the second part, 
\begin{talign*}
    \abs{H(x) - H(y)} &= \abs{\int^\infty_x z \dd F(z) - \int^\infty_y z \dd F(z)} \\
    &= \abs{\int^y_x z \dd F(z)} \\
    &= \abs{c \int^y_x \dd F(z)} \qq{for some $c\in[x,y]$} \\
    &\leq y_{\text{max}} \abs{\int^y_x \dd F(z)} = y_{\text{max}} \abs{F(y) - F(x)} \\
    &\leq y_{\text{max}} f_{\text{max}} \abs{x - y},
\end{talign*}
where we use the mean value theorem for integrals in the third inequality and~\eqref{eq:claim:HFLipschitz} in the final inequality.
\end{proof}

\subsection{Proof of Theorem~\ref{thm:optimal_convergence}}
\begin{proof}
Observe that $e^{-\epsilon} \geq 1 - \epsilon$.
Thus if $\frac{R^{(M)}}{r^\star} \geq e^{-\epsilon}$ then $\frac{R^{(M)}}{r^\star} \geq 1 - \epsilon$.
From Lemma~\ref{lem:reward_lower_bound} we have that 
\begin{equation}
  \textstyle\frac{R^{(M)}}{r^\star} \geq e^{-n(\delta + \overline{\lambda}\epsilon_{\overline{F}}T)} \qq{where} \delta = \max\qty{ \delta_{H}, \delta_{\overline{F}}}  
\end{equation} 
and $H(\widetilde{y}^{(M)}_k(t)) \geq e^{-\delta_H} H(y_k(t))$, $\overline{F}(\widetilde{y}^{(M)}_k(t)) \geq e^{-\delta_{\overline{F}}} \overline{F}(y_k(t))$ and $\abs*{\overline{F}(\widetilde{y}^{(M)}_k(t)) - \overline{F}(y_k(t))} \leq \epsilon_{\overline{F}}$ for all $k \in \qty{1, \ldots, n}$ and $t \in [0, T]$ and some $\delta_H, \delta_{\overline{F}} \in [0, 1]$, $\epsilon_{\overline{F}} \geq 0$. We thus have $\frac{R^{(M)}}{r^\star} \geq e^{-\epsilon}$ when $ \epsilon \geq n(\delta + \overline{\lambda}\epsilon_{\overline{F}}T)$.
In terms of probabilities, we have
\begin{talign}\label{eq:thm:optimal_convergence:1}
    &\mathbb{P}\qty[ \frac{R^{(M)}}{r^\star} \geq 1-\epsilon ] \geq \mathbb{P}\qty[ \epsilon \geq n(\delta + \overline{\lambda}\epsilon_{\overline{F}}T)] \nonumber \\
    &\quad\geq  \mathbb{P}\qty[\qty(\epsilon \geq \frac{n\delta}{2}) \wedge \qty(\epsilon \geq \frac{n \overline{\lambda}\epsilon_{\overline{F}}T}{2})] \nonumber \\
    &\quad\geq  \mathbb{P}\qty[\qty(\epsilon \geq \frac{n\delta_H}{2}) \wedge \qty(\epsilon \geq \frac{n\delta_{\overline{F}}}{2}) \wedge \qty(\epsilon \geq \frac{n \overline{\lambda}\epsilon_{\overline{F}}T}{2})] \\
    &\quad= 1 - \mathbb{P}\qty[\qty(\epsilon < \frac{n\delta_H}{2}) \vee \qty(\epsilon < \frac{n\delta_{\overline{F}}}{2}) \vee \qty(\epsilon < \frac{n \overline{\lambda}\epsilon_{\overline{F}}T}{2})] \nonumber \\
    &\quad\geq 1 - \mathbb{P}\qty[\epsilon < \frac{n\delta_H}{2}] - \mathbb{P}\qty[\epsilon < \frac{n\delta_{\overline{F}}}{2}] - \mathbb{P} \qty[\epsilon < \frac{n \overline{\lambda}\epsilon_{\overline{F}}T}{2}] \nonumber \\
    &\quad= 1 -  \mathbb{P}\qty[\delta_H > \frac{2\epsilon}{n}] - \mathbb{P}\qty[\delta_{\overline{F}} > \frac{2\epsilon}{n}] - \mathbb{P}\qty[\epsilon_{\overline{F}} > \frac{2\epsilon }{n \overline{\lambda}T}] \nonumber
\end{talign}
where we have used in the second, third and fourth lines that $A \Rightarrow B$ implies $\mathbb{P}(A) \leq \mathbb{P}(B)$ for events $A, B$.
We will now prove that as $M\to\infty$: \emph{i.} $\mathbb{P}\qty[\epsilon_{\overline{F}} > \frac{2\epsilon }{n \overline{\lambda}T}] \to 0$; \emph{ii.} $\mathbb{P}\qty[\delta_{\overline{F}} > \frac{2\epsilon}{n}] \to 0$; and \emph{iii.} $\mathbb{P}\qty[\delta_H > \frac{2\epsilon}{n}] \to 0$; from which the result follows upon substitution into~\eqref{eq:thm:optimal_convergence:1}.

\emph{Case i.} 
We have that 
\begin{talign}\label{eq:thm:optimal_convergence:1b}
&\epsilon_{\overline{F}} \leq \abs*{\overline{F}(\widetilde{y}^{(M)}_k(t)) - \overline{F}(y_k(t))} \nonumber \\
&\quad \leq f_{\text{max}}\abs*{\widetilde{y}^{(M)}_k(t) - y_k(t)} \qq{} (\text{\small Claim~\ref{claim:HFLipschitz}})  \nonumber \\ 
&\quad \leq f_{\text{max}} (\delta_\lambda + \delta_\phi) \mu \qty( e^{2 \lambda_{\text{max}} (T - t) } - 1 ) \qq{} (\text{\small Lemma~\ref{lem:threshold_dist_bound}}) \nonumber \\
&\quad \leq f_{\text{max}} (\delta_\lambda + \delta_\phi) \mu \qty( e^{2 \lambda_{\text{max}} T } - 1 ), 
\end{talign}
where $(1 - \delta_\lambda) \lambda(t) \leq \widetilde{\lambda}^{(M)}(t) \leq (1 + \delta_\lambda) \lambda(t)$ for all $t \in [0, T]$ and $(1 - \delta_\phi) \phi(y) \leq \widetilde{\phi}(y) \leq (1 + \delta_\phi) \phi(y)$ for all $y \in [0, y_{\text{max}}]$, and $0 < \delta_\lambda, \delta_\phi < 1$.
This leads to
\begin{talign}\label{eq:thm:optimal_convergence:2}
    \mathbb{P}\qty[\epsilon_{\overline{F}} > \frac{2\epsilon }{n \overline{\lambda}T}] & \leq \mathbb{P}\qty[ f_{\text{max}} (\delta_\lambda + \delta_\phi) \mu \qty( e^{2 \lambda_{\text{max}} T } - 1) > \frac{2\epsilon }{n \overline{\lambda}T} ] \nonumber \\
    &= \mathbb{P}\qty[ \delta_\lambda + \delta_\phi > \frac{2\epsilon }{n f_{\text{max}}  \overline{\lambda}T \mu \qty( e^{2 \lambda_{\text{max}} T } - 1)} ] \nonumber \\
    &= \mathbb{P}\qty[ \delta_\lambda + \delta_\phi > \frac{\nu}{\xi} ],
\end{talign} 
where, for brevity we write $\nu := \frac{2\epsilon}{n \overline{\lambda} T}$ and $\xi :=  f_{\text{max}}  \mu \qty( e^{2 \lambda_{\text{max}} T } - 1)$.
From~\eqref{eq:thm:optimal_convergence:2}, we have
\begin{talign}\label{eq:thm:optimal_convergence:3}
     &\mathbb{P}\qty[\epsilon_{\overline{F}} > \nu] \leq \mathbb{P}\qty[ \delta_\lambda + \delta_\phi > \frac{\nu}{\xi}] = 1 - \mathbb{P}\qty[ \delta_\lambda + \delta_\phi \leq \frac{\nu}{\xi}] \nonumber \\
    &\quad \leq 1 - \mathbb{P}\qty[ \qty(\delta_\lambda \leq \frac{\nu}{2\xi}) \wedge \qty(\delta_\phi \leq \frac{\nu}{2\xi})] \nonumber \\
     &\quad = 1 - \mathbb{P}\qty[ \delta_\lambda \leq \frac{\nu}{2\xi}] \mathbb{P}\qty[ \delta_\phi \leq \frac{\nu}{2\xi}] \qq{} \text{\small (independence)} \nonumber \\
     &\quad = 1 - \qty( 1 - \mathbb{P}\qty[ \delta_\lambda > \frac{\nu}{2\xi}] ) \qty( 1 - \mathbb{P}\qty[ \delta_\phi > \frac{\nu}{2\xi}] ) \nonumber \\
     &\quad = \mathbb{P}\qty[ \delta_\lambda > \frac{\nu}{2\xi}] + \mathbb{P}\qty[ \delta_\phi > \frac{\nu}{2\xi}] - \mathbb{P}\qty[ \delta_\lambda > \frac{\nu}{2\xi}]\mathbb{P}\qty[ \delta_\phi > \frac{\nu}{2\xi}] \nonumber \\
     &\quad \leq \mathbb{P}\qty[ \delta_\lambda > \frac{\nu}{2\xi}] + \mathbb{P}\qty[ \delta_\phi > \frac{\nu}{2\xi}].
\end{talign}
From the definition of $\delta_\phi$, we have for all $y \in [0, y_{\text{max}}]$, where $y_{\text{max}} \geq \widetilde{y}^{(M)}_k(t)$ and $y_{\text{max}} \geq y_k(t)$ for all $t \in [0, T]$, $k \in \{1, \ldots, n\}$, that
\begin{talign}\label{eq:thm:optimal_convergence:4}
    &\mathbb{P}[ \delta_\phi > \frac{\nu}{2\xi}] \leq \mathbb{P}[ (\widetilde{\phi}(y)> (1 + \frac{\nu}{2\xi}) \phi(y)) \nonumber\\
    &\quad\hspace{3.7cm} \vee (\widetilde{\phi}(y)  < (1 - \frac{\nu}{2\xi}) \phi(y)) ] \nonumber \\
    &\quad \leq \mathbb{P}[\widetilde{\phi}(y) > (1 + \frac{\nu}{2\xi}) \phi(y)]  \nonumber \\ &\quad\hspace{3.7cm} + \mathbb{P}[\widetilde{\phi}(y)  < (1 - \frac{\nu}{2\xi}) \phi(y)] \nonumber \\
    &\quad\leq \mathbb{P}[\phi(y) + \abs*{ \phi(y) - \widetilde{\phi}(y) }> (1 + \frac{\nu}{2\xi}) \phi(y)]  \nonumber \\
    &\quad\hspace{1.3cm} + \mathbb{P}[\phi(y) - \abs*{ \phi(y) - \widetilde{\phi}(y) } < (1 - \frac{\nu}{2\xi}) \phi(y)] \nonumber \\
    &\quad = \mathbb{P}[\abs*{ \phi(y) - \widetilde{\phi}_{N}(y) } > \frac{\nu}{2\xi} \phi(y)]  \nonumber \\ 
    &\quad\hspace{2.7cm} + \mathbb{P}[- \abs*{\phi(y)  - \widetilde{\phi}_{N}(y) } <  - \frac{\nu}{2\xi} \phi(y)] \nonumber \\
    &\quad = 2 \cdot \mathbb{P}[\abs*{ \phi(y) - \widetilde{\phi}_{N}(y) } > \frac{\nu}{2\xi} \phi(y)] 
\end{talign}
From Theorem~\ref{thm:ecdf_convergence}, we have for any $\zeta > 0$ that $\mathbb{P}[\abs*{ \phi(y) - \widetilde{\phi}(y) } > \zeta]$ goes to zero as $M \to \infty$, since as $M \to\infty$ the number of samples, $N$, used to compute $\widetilde{\phi}$ goes to infinity.
Furthermore, we have from Claim~\ref{claim:HFlowerbound} that $\phi(y) > 0$.
Setting $\zeta = \frac{\nu}{2\xi} \cdot  \min_{z \in [0, y_{\text{max}}]} \qty{\phi(z)}$, from~\eqref{eq:thm:optimal_convergence:4} we have that $\mathbb{P}[ \delta_\phi > \frac{\nu}{2\xi}] \to 0$ as $M \to 0$.
By a similar argument for $\delta_\lambda$, using Theorem~\ref{thm:rate_estimator_convergence} and the fact that $\lambda(t) > 0$ for all $t \in [0, T]$, we also have $\mathbb{P}[ \delta_\lambda > \frac{\nu}{2\xi}] \to 0$ as $M\to 0$. Substituting into~\eqref{eq:thm:optimal_convergence:3}, we find that $\mathbb{P}[\epsilon_{\overline{F}} > \frac{2\epsilon }{n \overline{\lambda}T}] \to 0$ as $M\to\infty$.

\emph{Case ii.}
We now show that $\mathbb{P}\qty[\delta_{\overline{F}} > \frac{2\epsilon}{n}] \to 0$ as $M \to \infty$.
By the monotonicity of $\exp(\,\cdot\,)$,
\begin{talign}
e^{-\delta_{\overline{F}}} & \leq \frac{\overline{F}(\widetilde{y}^{(M)}_k(t))}{\overline{F}(y_k(t))} \leq e^{\delta_{\overline{F}}} \Longleftrightarrow \nonumber \\
\delta_{\overline{F}} &\leq \abs{ \ln(\overline{F}(\widetilde{y}^{(M)}_k(t))) - \ln(\overline{F}(y_k(t))) } \nonumber \\
&\leq \frac{L_{\overline{F}}}{\beta_{\overline{F}}}\abs*{ \overline{F}(\widetilde{y}^{(M)}_k(t)) - \overline{F}(y_k(t)) } \nonumber \\
&\leq \frac{L_{\overline{F}} \xi}{\beta_{\overline{F}}}  (\delta_\phi + \delta_\lambda),  \qq{} \text{\small (from~\eqref{eq:thm:optimal_convergence:1b})} \nonumber
\end{talign}
where the penultimate line follows from Lemma~\ref{lem:lipschitz_prod}, Claim~\ref{claim:HFlowerbound} and Claim~\ref{claim:HFLipschitz}.
This yields
\begin{talign}\label{eq:thm:optimal_convergence:6}
\mathbb{P}\qty[\delta_{\overline{F}} > \frac{2\epsilon}{n}] &\leq \mathbb{P}\qty[\frac{L_{\overline{F}} \xi}{\beta_{\overline{F}}}  (\delta_\phi + \delta_\lambda) > \frac{2\epsilon}{n}] \nonumber \\
&= \mathbb{P}\qty[ \delta_\phi + \delta_\lambda > \frac{2\epsilon \beta_{\overline{F}}}{n L_{\overline{F}} \xi} ]  \\
&\leq \mathbb{P}\qty[ \delta_\phi > \frac{\epsilon \beta_{\overline{F}}}{n L_{\overline{F}} \xi} ] + \mathbb{P}\qty[ \delta_\lambda > \frac{\epsilon \beta_{\overline{F}}}{n L_{\overline{F}} \xi} ], \nonumber
\end{talign}
where the last line results from an argument identical to that in~\eqref{eq:thm:optimal_convergence:3}.
Using the same argument as in \emph{Case i.} following~\eqref{eq:thm:optimal_convergence:4}, we have that both terms in the final inequality~of~\eqref{eq:thm:optimal_convergence:6} go to zero as $M \to \infty$ and so $\mathbb{P}\qty[\delta_{\overline{F}} > \frac{2\epsilon}{n}] \to 0$, as required.

\emph{Case iii.} Finally, we verify that $\mathbb{P}\qty[\delta_H > \frac{2\epsilon}{n}] \to 0$ as $M \to \infty$ via an argument identical to \emph{Case ii.}, but replacing all instances of $\overline{F}$ with $H$. 

We have shown that the right hand side of the final inequality in~\eqref{eq:thm:optimal_convergence:1} approaches unity as $M \to \infty$ and the theorem is proved.
\end{proof}

\section{Experimental Details}

\paragraph{Data preprocessing.}

For both the \texttt{cc-fraud} and \texttt{ieee-fraud} datasets some basic data cleaning is conducted, namely columns where greater than 90\% of values are the same (including null) are removed.
The timestamp variable is replaced by two variables: the sine and cosine of the timestamps' difference with respect to the first occuring timestamp, normalised to a period of $T = \text{one calendar day}$.
The monetary value column is transformed according to a Yeo-Johnson transformation~\citeSM{yeo-johnson}.
Note that the monetary value is inverse transformed back for the NPSA algorithm validation.
Additionally for \texttt{ieee-fraud} the following are carried out: null values are imputed with the column median for numeric data or mode for categorical; and additional feature interactions from~\citeSM{ieee-processing-kernel} are introduced.

\paragraph{Model training.}

The \texttt{LGBMClassifier} from the LightGBM library~\citeSM{lightgbm} is used.
Cross validation is carried out via 60 iterations~\citeSM{random-search} of random search over the following hyperparameter space: 
\begin{lstlisting}[language=python]
{
    'num_leaves': [31, 50, 150, 500], 
    'min_data_in_leaf': [20, 100, 200],
    'bagging_fraction' : [0.1, 0.25, 0.9],
    'feature_fraction' : [0.1, 0.25, 0.9],
    'learning_rate': [0.01, 0.1, 0.3],
    'min_child_weight': [0.00001, 0.0001, 0.001, 0.01],   
    'reg_alpha': [1, 1.5, 2], 
    'reg_lambda': [1, 1.5, 2],
    'max_depth': [-1, 5, 25, 50]
}
\end{lstlisting}
The score used is the cross-validated $F_1$-score over 5 Time-series splits, where the splits are defined via \texttt{sklearn.preprocessing.TimeSeriesSplit}~\cite{scikit-learn}, and SMOTE~\citeSM{smote} with minority oversampling is used to address class imbalance.
The best hyperparameters are then used to retrain over the whole dataset to give the classifier \texttt{clf}.

\begin{figure}
    \centering
    \includegraphics[width=\columnwidth]{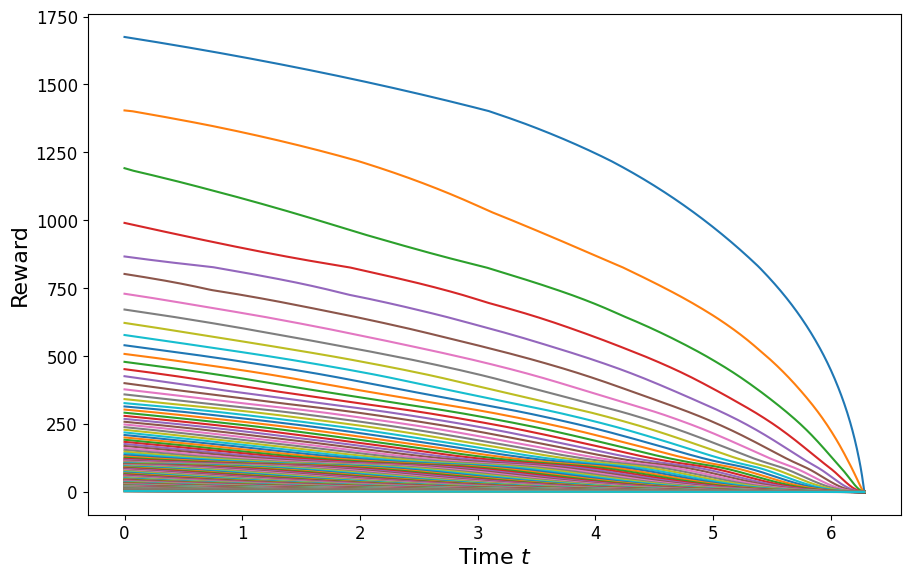}
    \includegraphics[width=\columnwidth]{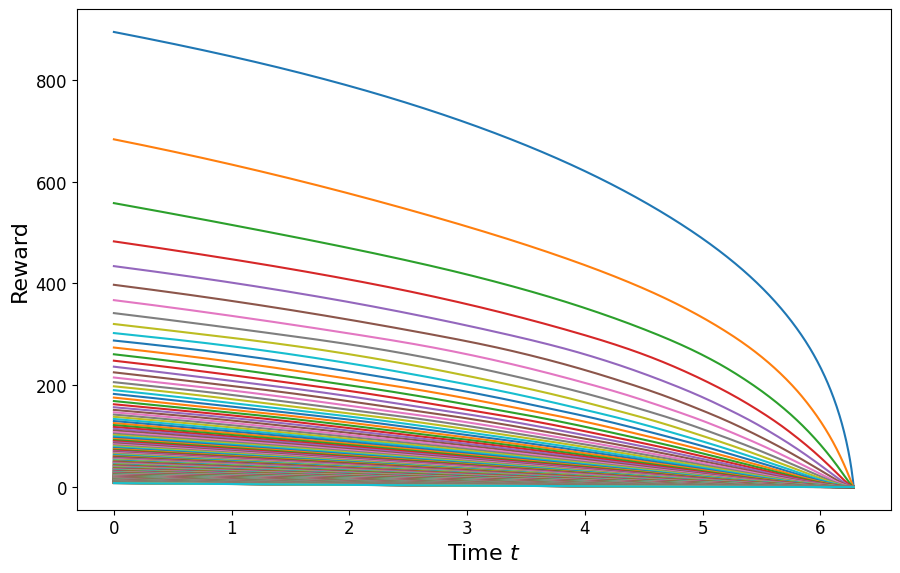}
    \caption{Derived threshold curves ($n=250$) for \texttt{cc-fraud} (top) and \texttt{ieee-fraud} (bottom) datasets.
    The time axis is scaled such that $T = 2\pi \approx 2.83$ represents one calendar day.}
    \label{fig:derived_thresholds}
\end{figure}

\paragraph{Computing critical curves.}

We learn $\widetilde{\phi}(y)$ via Algorithm~\ref{alg:ecdf_mean_shortage} where the $\{
x_i\}$ are the classifier confidence $\texttt{clf.predict\_proba(}\, \cdot \,\texttt{)}$ multiplied by monetary value of each transaction, evaluated over the training set.
The intensity $\widetilde{\lambda}(t)$ is evaluated as described in Section~\ref{sec:est_non_hom_Poisson}.

The solver $\mathcal{D}$ used in Algorithm~\ref{alg:main_meta} is \texttt{integrate.solve\_ivp} from scipy~\citeSM{2020SciPy-NMeth} with the default solver~\citeSM{DORMAND198019,Lawrence1986SomePR}, and accuracy hyperparameters \texttt{r\_tol = 1e-6}, \texttt{a\_tol = 1e-8}.

We provide concrete numbers on how much computation time is used for computing the critical curves in Table~\ref{tab:comptime}.

\begin{table}
\setlength\tabcolsep{3pt} 
\scriptsize
\centering
\begin{tabularx}{0.83\columnwidth}{lccc} 
\toprule
Dataset  & $\tau(\widetilde{\phi}_{\text{cache}})$ & $\tau(\widetilde{\phi}(y))$ & 
$\tau(\{y_k(t)\}_{k=1}^{250})$ \\
\midrule
\texttt{cc-fraud} & 1.030\,$\pm$\,0.02\,s & 3.310\,$\pm$\,0.004\,\textmu s & 28.3\,$\pm$\,0.1\,min \\
\addlinespace
\texttt{ieee-fraud} & 2.20\,$\pm$\,0.02\,s & 3.26\,$\pm$\,0.02\,\textmu s & 15.117\,$\pm$\,0.004\,min \\
\bottomrule
\end{tabularx}
\caption{
Wall-clock time taken $\tau(\,\cdot\,)$ to precompute the mean-shortage cache $\widetilde{\phi}_{\text{cache}}$, individual evaluations, $\widetilde{\phi}(y)$, and to fit 250 critical curves using NPSA, $\{y_k(t)\}_{k=1}^{250}$, on the two public fraud datasets used in experiments. 
All computation times measured on an AWS EC2 \texttt{c5n.9xlarge} instance and quantities are mean $\pm$ standard deviation of 7 runs.
}
\label{tab:comptime}
\end{table}

\paragraph{Fraud Experiment Thresholds.}

For reference we plot the derived NPSA threshold curves giving rise to the results reported in Figure~\ref{fig:expt:fraud_datasets} in Figure~\ref{fig:derived_thresholds}.

\bibliographystyleSM{named}
\bibliographySM{references}

\end{document}